\documentclass[11pt]{article}

\usepackage[preprint]{acl}

\usepackage{times}
\usepackage{latexsym}
\usepackage[T1]{fontenc}
\usepackage[utf8]{inputenc}
\usepackage{microtype}
\IfFileExists{inconsolata.sty}{\usepackage{inconsolata}}{}
\usepackage{graphicx}
\usepackage{amsmath,amssymb,amsthm,bm}
\usepackage{booktabs}
\usepackage{multirow}

\newtheorem{theorem}{Theorem}
\newtheorem{corollary}{Corollary}
\newtheorem{proposition}{Proposition}
\newtheorem{assumption}{Assumption}

\newcommand{\R}{\mathbb{R}}
\newcommand{\E}{\mathbb{E}}

\newcommand{\Cov}{\operatorname{Cov}}
\newcommand{\tr}{\operatorname{tr}}
\newcommand{\bias}{\operatorname{Bias}}
\newcommand{\MSE}{\operatorname{MSE}}
\newcommand{\mllm}{\bm m_{\mathrm{LLM}}}
\newcommand{\claude}{\texttt{claude-opus-4-7}}
\newcommand{\gemini}{\texttt{gemini-3.1-flash-lite}}
\newcommand{\geminew}{\texttt{gemini-3-flash-preview}}

\title{When Do LLM-Elicited Priors Help? A Controlled Audit of Language
Models as Priors for Small-Sample Bayesian Regression}

\author{Ran Zou\thanks{Equal contribution.} \\
  Department of Statistics \\
  University of California, Irvine \\
  \texttt{aprilz3@uci.edu} \And
  Yang Liu\footnotemark[1] \\
  Department of Statistics and Operations Research \\
  University of North Carolina at Chapel Hill \\
  \texttt{yangliu1@unc.edu}}

\begin{document}
\maketitle
\begingroup\renewcommand{\thefootnote}{}%
\footnotetext{\raggedright Code and data: \url{https://github.com/AprilZou852/llm-elicited-priors}}%
\endgroup

\begin{abstract}
Large language models (LLMs) are increasingly asked where to center
the coefficients of a small-sample Bayesian regression, but the
evidence is unclear: benchmarks are often built to match the prior
and models may remember them.  We audit this practice with one
frontier and one flash-tier model, using hidden feature names,
post-cutoff housing data and a pre-specified test that gives
features the wrong descriptions.  First, the models share their
errors (correlation $0.92$--$0.99$ across draws, vendors and
paraphrases): averaging $20$ draws is no more accurate than one, and
treating draws as independent creates overconfidence.  Much of this
error comes from reporting each feature's effect alone, not with the
others held fixed.  Second, against the best classical prior, the
LLM center helps clearly on one of four benchmarks (RMSE $-5.8\%$),
marginally on two, and hurts mean RMSE on the fourth.  Third, the
LLM's stated uncertainty is unreliable; letting the data set the
prior scale cuts the harm of a wrong center four-fold against a
unit-variance prior.  A pre-specified external test returned the
verdict \textsc{partial}: on $18$ rule-selected problems the center
helps on six and hurts on four, and a hard trust gate trusts a
harmful prior while an evidence-weighted mixture is safer.
\end{abstract}

\section{Introduction}\label{sec:intro}

\begin{figure*}[t]\centering
\includegraphics[width=\textwidth]{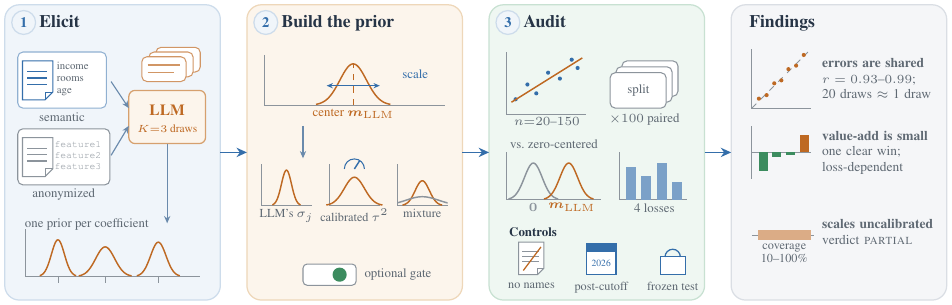}
\caption{Overview of the audit.  (1)~A fixed prompt, with real feature
names or with names removed, elicits a prior mean and scale per
coefficient from the LLM.  (2)~The elicited mean becomes the prior
center; its scale is the LLM's own, calibrated on the data
(Theorem~\ref{thm:cal}), or a robust mixture.  (3)~The priors are compared
with zero-centered adaptive priors on 100 paired small-sample splits,
with memorization controls, an error-dependence measurement and a
pre-specified external test.  }
\label{fig:overview}
\end{figure*}

Specifying informative priors is a perennial difficulty in applied
Bayesian analysis: defaults are either so broad that the posterior
behaves like maximum likelihood, or so narrow that they bias
inference in ways that are hard to audit
\citep{ohagan2006,garthwaite2005}.  A recent line of work proposes
an appealing shortcut: describe the problem and features to a large
language model (LLM) and let it return Gaussian hyperparameters
$(\mu_j,\sigma_j^2)$ per standardized coefficient.
\citet{selby2025} established that a zero-shot LLM can return usable
quantitative priors; AutoElicit \citep{capstick2025} builds a mixture
prior from many rephrasings of the prompt, which we summarize by
moment matching; \citet{gouk2024} elicit priors through LLM-generated
synthetic data; \citet{riegler2025} identify scale calibration as
the central open difficulty; and the recipe is spreading, to
survival analysis \citep{evans2025glioblastoma}, to automated
Bayesian pipelines \citep{huang2025llmbi} and to multi-agent
schemes that aggregate agents' priors by logarithmic pooling
\citep{huang2025llmprior}.

The evidence for this recipe is less clear than the headline numbers
suggest.  \emph{First, the ground truth is often built to match the
prior}: semi-synthetic benchmarks set $\bm\beta^\star$ close to what
an expert, or the LLM itself, would propose, so the displacement
$\delta=\lVert\bm m_0-\bm\beta^\star\rVert$ is small by
construction.  \emph{Second, memorization is hard to rule out}: LLMs
can reproduce training data \citep{carlini2023,nasr2023}, benchmark
contamination is a known risk \citep{magar2022,oren2024}, and standard
benchmarks appear with coefficient signs in hundreds of tutorials,
so an LLM may be remembering rather than reasoning.  AutoElicit
does test for memorized rows and adds a private dataset.  \emph{Third,
the LLM, not the data, sets the prior scale}: AutoElicit takes it
from the LLM's own elicited scales, yet nominal $95\%$ intervals
elicited from eleven LLMs cover the truth only $9$--$44\%$ of the
time \citep{hobor2026elicitation}, and across six frontier models
confidence is only weakly related to accuracy
\citep{marzoev2026openestimate}.  Calibration of model confidence is
a long-standing concern \citep{guo2017,lin2022,kadavath2022,tian2023};
we measure the downstream cost of a miscalibrated scale.  \emph{Fourth, dependence between
draws is not measured}: AutoElicit mixes over many rephrasings of
one prompt, and the multi-agent schemes log-pool across agents.  Linear and logarithmic pooling are
classical \citep{stone1961,french1985,bordley1982,genest1986,ranjan2010},
but dependence between sources is known to limit what combining
them can gain \citep{winkler1981,clemen1985}.  To our knowledge, no prior
study compares an elicited center with zero-centered adaptive priors while
controlling for contamination, or measures the dependence that
aggregation ignores.

\emph{This paper} treats the elicited mean as an input of unknown
quality and lets the data decide how much to trust it: \emph{how
should such a mean, whose stated uncertainty is unreliable, enter a
Bayesian regression so that it helps when it is accurate and its
harm is measured and limited when it is not?}  The tools are
classical. One is type-II maximum
likelihood for Gaussian priors
\citep{mackay1992,tipping2001,wipf2008,bishop2006}, which we apply
around a non-zero elicited mean.  The other is borrowing from
historical data \citep{viele2014}: power priors \citep{ibrahim2000}, commensurate
priors \citep{hobbs2011}, of which our trust dial is an instance,
robust meta-analytic-predictive mixtures
\citep{schmidli2014}, of which our robust mixture is one,
and biased-target and multiple shrinkage
\citep{green1991,george1986}. 
Figure~\ref{fig:overview} summarizes the audit.

Our contributions are:
\textbf{(1)~An audit that controls for contamination.}  A fully
recorded two-vendor elicitation on six benchmarks, including
post-cutoff NYC-2026 housing data and a stress test whose random
$\bm\beta^\star$ was fixed before elicitation, with contamination
controls, strong zero-centered adaptive baselines and one shared
split grid.  \claude{} contributes $K{=}3$ draws on five of the six benchmarks;
\gemini{} runs every LLM-specific battery.
\textbf{(2)~Measurements of dependence and aggregation.}
Elicitation errors correlate $0.93$--$0.99$ within and across
vendors and $0.92$--$0.99$ across paraphrases
(Section~\ref{sec:E4}), and averaging $20$ real draws is no more
accurate than using one (Figure~\ref{fig:e19}).  Repeated or
rephrased sampling adds no independent evidence, so aggregators that
sum precisions become overconfident
(Theorem~\ref{thm:logpool}).  Much of the shared error comes from
marginal-effect reporting.
\textbf{(3)~A single-protocol comparison with adaptive classical
priors.}  Tables~\ref{tab:rmse} and~\ref{tab:losses} compare the main
methods on one split grid under several losses, with the identities
of Section~\ref{sec:theory} as reference.  The gain over the best
zero-centered adaptive prior is small and depends on the loss: clear
on NYC, marginal on Diabetes, only at $n{=}20$ on semi-synthetic
California, and negative in mean RMSE on real California.  On three
of four benchmarks the center is within a few percent of an oracle
that knows the population marginal slopes.
\textbf{(4)~Harm accounting.}  Elicited scales are unreliable, with
coverage from $0.10$ to $1.00$.  A type-II ML trust dial cuts the
harm of a wrong center four-fold against a weak
$\mathcal N(\bm0,\bm I)$ prior and leaves none against a wider one.
A pre-specified external validation on six rule-selected datasets
returned \textsc{partial} because the hard gate trusts one harmful
prior.  With a pre-specified extension to $18$ external problems, the center
helps on six and hurts on four, while the evidence-weighted mixture
keeps relative regret within $7.5\%$ on every external cell.

\subsection{Related Work}\label{sec:related}

Expert elicitation has a long literature
\citep{ohagan2006,garthwaite2005,kadane1980,kadane1998} and
audit-trail protocols such as SHELF \citep{oakley2010}; because
experts can be over-confident \citep{ohagan2006,kynn2008},
Theorem~\ref{thm:cal} asks only for a mean.  Zero-centered shrinkage priors include
ridge \citep{hoerl1970}, $g$-priors \citep{zellner1986,liang2008},
spike-and-slab \citep{mitchell1988,george1993}, the Bayesian lasso
\citep{park2008} and global-local priors
\citep{carvalho2010,piironen2017,makalic2016}; we run
cross-validated ridge, empirical Bayes and the horseshoe, and
per-coordinate empirical Bayes beats the LLM prior in mean RMSE on
real California.  Related but distinct are implicit-prior
elicitation \citep{zhu2024eliciting} and prior-fitted or
text-conditioned predictors
\citep{muller2022pfn,hollmann2023,requeima2024}, which return
predictions rather than a coefficient posterior.

\section{Setup and Notation}\label{sec:setup}

$\bm y\in\R^n$ is the response, $\bm X\in\R^{n\times p}$ the design of
standardized features.  We use the conjugate Gaussian model
\[
\bm y\mid\bm X,\bm\beta,\sigma^2\sim\mathcal N(\bm X\bm\beta,\sigma^2\bm I),
\;\;\bm\beta\sim\mathcal N(\bm m_0,\bm S_0),
\]
whose posterior is Gaussian \citep[\S3.3]{bishop2006}:
\begin{align}
\bm S_n&=(\bm S_0^{-1}+\sigma^{-2}\bm X^\top\bm X)^{-1},\label{eq:posterior}\\
\bm m_n&=\bm S_n(\bm S_0^{-1}\bm m_0+\sigma^{-2}\bm X^\top\bm y),\nonumber
\end{align}
with predictive
$\mathcal N(\bm x_\star^\top\bm m_n,\,\sigma^2+\bm x_\star^\top\bm S_n\bm x_\star)$
at a test point $\bm x_\star$.  Let
$\bm\delta:=\bm m_0-\bm\beta^\star$ and $\delta:=\lVert\bm\delta\rVert$.

\begin{assumption}[Regularity]\label{ass:reg}
$\bm X^\top\bm X\succ0$, $\bm S_0\succ0$, $\sigma^2>0$,
$\bm y\mid\bm X\sim\mathcal N(\bm X\bm\beta^\star,\sigma^2\bm I)$;
expectations condition on $\bm X$.
\end{assumption}
\noindent Estimating $\sigma^2$ per prior would let the noise scale
depend on the prior, so our default fixes $\sigma^2=\sigma_\star^2$
across priors (an oracle for the noise scale only; checks with
estimated $\sigma^2$: Appendix~\ref{app:addl}).

\section{Theory}\label{sec:theory}

These are classical identities, stated so that the audit has exact
reference quantities.

\subsection{Risk Accounting: Bias--Variance of \texorpdfstring{$\bm m_n$}{the Posterior Mean}}

\begin{theorem}[Bias--variance of $\bm m_n$]\label{thm:bv}
Under Assumption~\ref{ass:reg} with $\sigma^2$ known,
$\bias(\bm m_n)=\bm S_n\bm S_0^{-1}\bm\delta$,
$\Cov(\bm m_n)=\bm S_n-\bm S_n\bm S_0^{-1}\bm S_n$, and
\begin{equation}\label{eq:mse}
\MSE(\bm m_n)=\lVert\bm S_n\bm S_0^{-1}\bm\delta\rVert^2+\tr(\bm S_n-\bm S_n\bm S_0^{-1}\bm S_n).
\end{equation}
\end{theorem}

\noindent\emph{(Proof in Appendix~\ref{app:thm1}.)}  An informative
prior beats the flat one iff its variance saving exceeds its squared
bias.

\begin{corollary}[Orthogonal crossover]\label{cor:cross}
If $\bm X^\top\bm X=n\bm I$ and $\bm S_0=\tau^2\bm I$, the
coordinate-wise MSE is
\begin{equation}\label{eq:msej}
\MSE_j(n)=\frac{\sigma^4\delta_j^2+n\sigma^2\tau^4}{(\sigma^2+n\tau^2)^2},
\end{equation}
versus $\sigma^2/n$ in the flat limit, so the informative prior wins
on coordinate $j$ iff $n(\delta_j^2-2\tau^2)\le\sigma^2$: it
dominates at every $n$ if $\delta_j^2\le2\tau^2$, else up to the
crossover $n^\star_j=\sigma^2/(\delta_j^2-2\tau^2)$ (proof:
Appendix~\ref{app:cor}).
\end{corollary}

$n^\star_j$ depends on the unknown $\bm\beta^\star$; the estimable
surrogate used below is the calibrated scale $\tau^{\star2}$ of
Theorem~\ref{thm:cal}.

\subsection{Trust Calibration: Type-II ML Around \texorpdfstring{$\mllm$}{a Non-Zero Mean}}\label{sec:thm2}

Given an elicited mean $\mllm$, treat the prior precision
$\alpha=1/\tau^2$ as a hyperparameter chosen by maximizing the log
marginal likelihood
\begin{multline}\label{eq:marglik}
\log p(\bm y\mid\mllm,\alpha,\sigma^2)
=-\tfrac12\big[n\log(2\pi)\\
+\log|\bm A_\alpha|+\bm r^\top\bm A_\alpha^{-1}\bm r\big],
\quad\bm r:=\bm y-\bm X\mllm,\\
\bm A_\alpha:=\sigma^2\bm I+\tfrac1\alpha\bm X\bm X^\top.
\end{multline}

\begin{theorem}[Fixed-point characterization of $\alpha^\star$]\label{thm:cal}
Under Assumption~\ref{ass:reg}, any stationary point
$\alpha^\star\in(0,\infty)$ of \eqref{eq:marglik} satisfies
\begin{equation}\label{eq:alphastar}
\alpha^\star=\gamma(\alpha^\star)\big/
\lVert\bm m_{\mathrm{MAP}}(\alpha^\star)-\mllm\rVert^2,
\end{equation}
with $\bm m_{\mathrm{MAP}}(\alpha)=(\bm X^\top\bm X/\sigma^2+\alpha\bm I)^{-1}
(\bm X^\top\bm y/\sigma^2+\alpha\mllm)$ and
$\gamma(\alpha)=\sum_{j}\lambda_j/(\lambda_j+\alpha\sigma^2)$,
$\lambda_j$ the eigenvalues of $\bm X^\top\bm X$.  A monotone EM
update and the MacKay iteration share these stationary points, and
the latter converges in nearly all our splits (proof, EM derivation, non-concavity:
Appendix~\ref{app:thm2}).
\end{theorem}

\paragraph{Attribution and guarantee.}  Subtracting the elicited
mean, $\tilde{\bm y}:=\bm y-\bm X\mllm$, turns Theorem~\ref{thm:cal}
into standard zero-centered evidence maximization
\citep{mackay1992,bishop2006}; likewise the ARD variant
\citep{tipping2001} and the GLM update of
Proposition~\ref{thm:logcal}.  Here it serves only as a trust dial
for elicited means whose own scales cannot be trusted.  In the
orthogonal design it is James--Stein shrinkage toward $\mllm$, which
for $p\ge5$ beats flat/OLS for every $\bm\beta^\star$
\citep{james1961,efron1973} (Proposition~\ref{prop:adaptive}).  This
no-harm guarantee is \emph{against the flat prior only}, not against
the weak or EB comparators (Section~\ref{sec:E5rev}); a uniform
oracle inequality \citep{stein1981,xie2012} remains open.

\subsection{Aggregating \texorpdfstring{$K$}{K} Elicitations}\label{sec:thm3}

Multi-sample recipes draw $K$ elicitations $(\bm m_k,\bm S_k)_{k=1}^K$
and summarize them.  Moment matching (the Gaussian summary of the
AutoElicit mixture) gives
\begin{equation}\label{eq:mm}
\begin{aligned}
\bm m_{\mathrm{mm}}&=\tfrac1K\textstyle\sum_k\bm m_k,\\
\bm S_{\mathrm{mm}}&=\tfrac1K\textstyle\sum_k\bm S_k+\widehat\Cov(\{\bm m_k\}),
\end{aligned}
\end{equation}
while the logarithmic pool \citep{genest1986} with unit weights gives
\begin{equation}\label{eq:lp}
\bm S_{\mathrm{lp}}^{-1}=\textstyle\sum_k\bm S_k^{-1},\qquad
\bm m_{\mathrm{lp}}=\bm S_{\mathrm{lp}}\textstyle\sum_k\bm S_k^{-1}\bm m_k.
\end{equation}

\begin{theorem}[Log-pooling; Appendix~\ref{app:thm3}]\label{thm:logpool}
If the $\bm m_k$ are $\bm\beta$ plus independent $\mathcal
N(\bm0,\bm S_k)$ errors and $\bm\beta$ has a flat prior, the log pool
\eqref{eq:lp} is the exact posterior of $\bm\beta$, with covariance
$\bm S/K$ for equal $\bm S_k=\bm S$, strictly below
$\bm S_{\mathrm{mm}}$ for $K\ge2$; if the errors share a fixed bias
$\bm b\neq\bm0$, the pooled mean converges to $\bm\beta+\bm b$ while
the covariance still shrinks as $1/K$, so the pool becomes confidently
wrong: its credible regions cover $\bm\beta$ with probability tending
to zero.
\end{theorem}

Draws from a \emph{single} model share its systematic errors, so
log-pooling is Bayes-optimal only under conditional independence,
whose violation Section~\ref{sec:E4} measures ($0.93$--$0.99$).  A
Laplace/MacKay analogue (Proposition~\ref{thm:logcal},
Appendix~\ref{app:thm4}; an approximation) carries the dial to
logistic regression; E9 (Appendix~\ref{app:theoryexp}) tests it.

\section{Experiments}\label{sec:exp}

\begin{table*}[t]\centering\footnotesize
\caption{Held-out RMSE at $n{=}20$ and $50$ (100 paired splits).  Zero-centered: Weak $\mathcal N(\bm0,\bm I)$, EB-iso, EB-pc (ARD).  LLM-centered: LLM-mom, the moment match of the draws (1v $\approx$ LLM-fixed in the text); LLM-cal, scale set by the data (Theorem~\ref{thm:cal}); LLM-pc, the same per coordinate.  1v: \claude{} ($K{=}3$); 2v: plus \gemini{} ($K{=}6$); NYC-2026: \gemini{} only.  \textbf{Bold}: best in row.  $^\ast$: beats the row's best zero-centered prior (paired Wilcoxon $p<0.05$).  Real California reverses under other losses (Table~\ref{tab:losses}).  $^\ddagger$TabPFN \citep{hollmann2025}: prediction only, not ranked.}
\label{tab:rmse}
\setlength{\tabcolsep}{6pt}
\begin{tabular}{llcccccccc}
\toprule
Benchmark & $n$ & Weak & EB-iso & EB-pc & LLM-mom (1v) & LLM-cal & LLM-pc & LLM-mom (2v) & TabPFN$^\ddagger$ \\
\midrule
Calif.\ (semi) & 20 & 0.714 & 0.730 & 0.715 & 0.691$^\ast$ & 0.708 & 0.712 & \textbf{0.690} & 0.756 \\
Calif.\ (semi) & 50 & \textbf{0.580} & 0.585 & 0.583 & 0.583 & 0.599 & 0.592 & 0.590 & 0.599 \\
Diabetes & 20 & 0.972 & 0.852 & 0.893 & 0.846 & 0.850 & 0.897 & \textbf{0.834}$^\ast$ & 0.878 \\
Diabetes & 50 & 0.799 & 0.772 & 0.787 & 0.771 & 0.766$^\ast$ & 0.785 & \textbf{0.766}$^\ast$ & 0.785 \\
NYC-2026 & 20 & 1.125 & 1.074 & 1.085 & \textbf{1.011}$^\ast$ & 1.019$^\ast$ & 1.050$^\ast$ & 1.011$^\ast$ & 1.085 \\
NYC-2026 & 50 & 0.906 & 0.900 & 0.895 & 0.876$^\ast$ & \textbf{0.870}$^\ast$ & 0.885$^\ast$ & 0.876$^\ast$ & 0.894 \\
Calif.\ (real) & 20 & 2.193 & 2.071 & \textbf{1.726} & 2.162 & 2.171 & 2.163 & 2.005 & 0.817 \\
Calif.\ (real) & 50 & 1.887 & 1.865 & \textbf{1.673} & 1.898 & 1.925 & 1.887 & 1.848 & 0.632 \\
\bottomrule
\end{tabular}
\end{table*}

Appendix~\ref{app:index} indexes every experiment code.

\paragraph{Elicitation protocol and provenance.}\label{sec:elicit}
One fixed prompt (Appendix~\ref{app:prompt}) asks the model, as an
expert statistician, to give each standardized coefficient a prior
mean $\mu_j$, a standard deviation $\sigma_j$ and a one-sentence
reason, ending in machine-readable JSON.  We call $\sigma_j$ the
\emph{elicited scale}: it is the model's own uncertainty, unlike the
\emph{calibrated scale} $\tau^{\star2}$ of Theorem~\ref{thm:cal}, which the
data choose.  Two vendors answer it.  \claude{} (Anthropic) is the primary
elicitor: $K{=}3$ recorded responses to the same prompt on every
benchmark except NYC-2026.  \gemini{} (Google) repeats this elicitation
($K{=}3$ each) and is the only elicitor for NYC-2026, its probe, the
$K{=}20$ draws, the partial-effect re-elicitation and every battery
(see Limitations).  We write \emph{1v} for \claude{} alone ($K{=}3$)
and \emph{2v} for the pooled draws of both vendors ($K{=}6$).  All
prompts, responses and parsed $(\mu_j,\sigma_j)$ ship verbatim with
the code and reproduce offline.  Every dataset is elicited twice: a
\emph{semantic} prior with the real feature names, and an
\emph{anonymized} one with the names replaced by
\texttt{feature\_1,\dots,feature\_p} and the description replaced by a content-free sentence,
which is our contamination control.
\emph{Datasets:} (1)~\emph{California, semi-synthetic}: $p{=}8$
features with the real California Housing correlation structure and
$\bm\beta^\star$ fixed near the real-data OLS fit
\citep{pace1997} before any elicitation; (2)~\emph{California, real}
\citep{pace1997}, $\bm\beta^\star$ $=$ full-data OLS;
(3)~\emph{Diabetes} \citep{efron2004}, $n{=}442$, $p{=}10$;
(4)~\emph{Breast Cancer (Wisconsin)} \citep{street1993}, a real binary task, $p{=}10$;
(5)~a \emph{logistic stress test} with random
$\bm\beta^\star\sim\mathcal N(\bm0,0.36\bm I)$ drawn from a seeded
RNG before elicitation, $p{=}6$; (6)~NYC-2026
(Section~\ref{sec:E18}).

\paragraph{Methods and protocol.}  Baseline Methods: flat
$\mathcal N(\bm0,10^2\bm I)$, weak $\mathcal N(\bm0,\bm I)$, CV ridge,
isotropic and per-coordinate empirical Bayes (EB-iso, EB-pc/ARD)
\citep{mackay1992,tipping2001}, and the horseshoe \citep{makalic2016}.  LLM
Methods: LLM-mom (1v) is the moment match of the \claude{} draws, and
LLM-fixed is the same prior with a diagonal covariance.  LLM-mom (2v)
and LLM-pool (2v) are the moment match and the log pool of both
vendors' draws.  LLM-cal is the calibrated prior of
Theorem~\ref{thm:cal}, LLM-mix the robust mixture and LLM-anon the
anonymized control.
\textbf{Every method in Tables~\ref{tab:rmse}--\ref{tab:losses} and
Appendix~\ref{app:t1} uses the same seeded grid of $100$ paired random
splits per (benchmark, $n$)}; Tier-2 cells use $40$ paired splits and
the stress family $50$, and the R-tables state their own counts.  Standardization is fit on the
training rows only. Every headline percentage on this grid is
a paired comparison with the best zero-centered baseline for that
benchmark and loss: weak, EB-iso, EB-pc or CV ridge.  Each has a
$95\%$ paired-bootstrap CI and a paired Wilcoxon test.  Comparisons
with weak are secondary.  Stars mark raw $p<0.05$.  A
Benjamini--Hochberg correction within each benchmark, $n$ and loss
would remove only two stars: semi-synthetic California RMSE at
$n{=}20$ with $q=0.052$, and real-California trimmed RMSE for LLM-mom
(2v) with $q=0.08$.  A $p$-value covers the random splits of one
dataset only.  Across problems the sample is the six problems of
Table~\ref{tab:t1cross}.

\subsection{When Does the LLM Prior Help?}\label{sec:E1}
\begin{table*}[t]\centering\footnotesize
\caption{Three losses at $n{=}20$, with splits and elicitors as in Table~\ref{tab:rmse}.  RMSE$_{\mathrm{trim}}$: RMSE on the test points within the 99th percentile of the training points' Mahalanobis distances.  Coef.\ err.: $\|\bm m_n-\bm\beta^\star\|$.  \textbf{Bold}: best of the six priors.  $^\ast$: beats the best zero-centered prior for that loss (paired Wilcoxon $p<0.05$).  $n{=}35$ and $50$: Appendix~\ref{app:t1}.}
\label{tab:losses}
\setlength{\tabcolsep}{4pt}
\resizebox{\textwidth}{!}{%
\begin{tabular}{lcccccccccccc}
\toprule
 & \multicolumn{3}{c}{Calif.\ (semi)} & \multicolumn{3}{c}{Diabetes} & \multicolumn{3}{c}{NYC-2026} & \multicolumn{3}{c}{Calif.\ (real)} \\
\cmidrule(lr){2-4}\cmidrule(lr){5-7}\cmidrule(lr){8-10}\cmidrule(lr){11-13}
Method & RMSE & RMSE$_{\mathrm{trim}}$ & coef.\ err. & RMSE & RMSE$_{\mathrm{trim}}$ & coef.\ err. & RMSE & RMSE$_{\mathrm{trim}}$ & coef.\ err. & RMSE & RMSE$_{\mathrm{trim}}$ & coef.\ err. \\
\midrule
Weak $\mathcal N(\bm0,\bm I)$ & 0.714 & 0.654 & \textbf{0.684} & 0.972 & 0.887 & 0.983 & 1.125 & 0.996 & 0.691 & 2.193 & 0.705 & \textbf{0.753} \\
EB-iso & 0.730 & 0.665 & 0.888 & 0.852 & 0.839 & 0.747 & 1.074 & 0.956 & 0.531 & 2.071 & 0.731 & 0.978 \\
EB-pc (ARD) & 0.715 & 0.657 & 0.933 & 0.893 & 0.852 & 0.808 & 1.085 & 0.972 & 0.568 & \textbf{1.726} & 0.722 & 1.072 \\
LLM-mom (1v) & 0.691$^\ast$ & \textbf{0.646}$^\ast$ & 0.809 & 0.846 & 0.833$^\ast$ & \textbf{0.701}$^\ast$ & \textbf{1.011}$^\ast$ & \textbf{0.954} & \textbf{0.401}$^\ast$ & 2.162 & \textbf{0.686}$^\ast$ & 0.796 \\
LLM-mom (2v) & \textbf{0.690} & 0.647 & 0.885 & \textbf{0.834}$^\ast$ & \textbf{0.824}$^\ast$ & 0.705$^\ast$ & 1.011$^\ast$ & 0.954 & 0.401$^\ast$ & 2.005 & 0.689$^\ast$ & 0.835 \\
LLM-cal & 0.708 & 0.659 & 0.979 & 0.850 & 0.835$^\ast$ & 0.703$^\ast$ & 1.019$^\ast$ & 0.956 & 0.414$^\ast$ & 2.171 & 0.707 & 0.923 \\
\midrule
TabPFN (pred.\ only) & 0.756 & 0.683 & --- & 0.878 & 0.856 & --- & 1.085 & 0.962 & --- & 0.817 & 0.724 & --- \\
\bottomrule
\end{tabular}}
\end{table*}

Table~\ref{tab:rmse} gives the main result and Table~\ref{tab:losses}
shows how it changes with the loss.

The \claude{} prior helps at
$n{=}20$ on semi-synthetic California, where it lowers RMSE by $3.2\%$ against weak, the best
classical method there.  From $n{=}35$ on no LLM method beats weak,
and the elicited mean has two wrong signs, with $\delta\approx1.10$.
On Diabetes the \claude{} prior changes RMSE by $-0.8\%$ and
coefficient error by $-6.3\%$ against EB-iso; pooling both vendors'
draws gives $-2.2\%$ in RMSE.  Real California shows the opposite in
mean RMSE.  EB-pc beats every single-vendor LLM prior at every $n$,
$1.726$ against $2.160$ at $n{=}20$, and not just by shrinking
harder: $\mathcal N(\bm0,0.1\bm I)$ gives $2.039$.  Its advantage
comes from a few extreme test points, as does that of the two-vendor
log pool, which reaches $1.093$ (Table~\ref{tab:t1realca}).  On the
test points within the range of the training data (RMSE$_{\mathrm{trim}}$) the
order reverses: the \claude{} prior is best at $0.686$ and EB-pc has
$0.722$.  The LLM priors are also best in median error.  In
coefficient error weak is best at $0.753$ and EB-pc the worst of the
six priors at $1.072$, and EB-pc's $90\%$ intervals cover only $31\%$
of the coefficients (Table~\ref{tab:t1coefcov}).  So on heavy-tailed
real data no method wins under every loss.

Is the LLM itself the active ingredient?
Table~\ref{tab:t1centers} gives every candidate center the same
calibrated scale.  Marginal slopes estimated from the $n{=}20$ sample
are too noisy: $7$--$13\%$ worse than the elicited center on every
benchmark.  The population marginal slopes, an oracle, are worse than
the elicited center on Diabetes, tied on semi-synthetic California,
and better on NYC and real California.  So the center carries roughly
population-level marginal knowledge at no cost in data, and on
collinear designs a $\delta$ scored against partial effects
overstates its error.  Neither prompting (Table~\ref{tab:R22}) nor
algebra repairs this estimand mismatch: converting the center to
partial effects with the full-data correlation matrix hurts
everywhere.  Finally, on real Breast Cancer the \claude{} prior
lowers the $n{=}20$ log-loss from $0.279$ to $0.247$, a gain gone by
$n{=}80$; the anonymized prior matches weak and the \gemini{} prior
harms (Appendix~\ref{app:crossmodel}).

\subsection{Failure Modes, Controls, Replication}\label{sec:E4}

Without feature names, \claude{} returns $\mathcal N(\bm0,\bm I)$ on
every draw, exactly the weak prior (Figure~\ref{fig:e4}).  \gemini{}
returns zero or near-zero means on all $28$ problems, with standard
deviations mostly between $0.3$ and $0.5$.  So the benefit comes from
the names: without them the model declines to guess.  Forbidding the
zero-mean default uncovers no hidden knowledge: the forced prior is a
fixed positional template with chance-level sign agreement, $0.52$,
and loses to the best classical prior on $20$ of $26$ regression
problems (Appendix~\ref{app:t2}).  This control cannot tell
understanding from recall; Section~\ref{sec:E18} does.

\begin{figure*}[t]\centering
\includegraphics[width=0.8\textwidth]{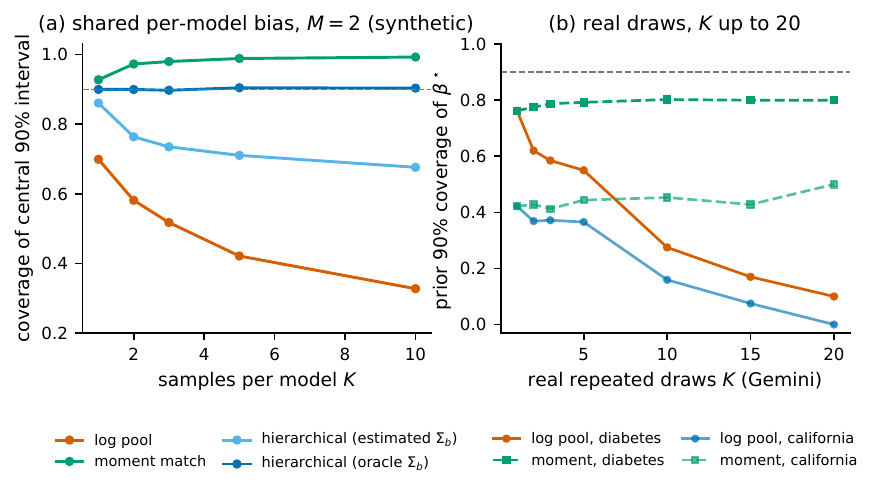}
\caption{E19 (\gemini): pooling draws as if they were independent makes the prior overconfident.  (a)~Simulation with a shared bias: the log pool's nominal-$90\%$ coverage of $\bm\beta$ collapses as $K$ grows, the oracle hierarchical pool keeps it, and moment matching stays conservative.  (b)~Real draws up to $K{=}20$: the same collapse, while moment matching stays flat.  The averaged center is no closer to the truth at $K{=}20$ than at $K{=}1$.}
\label{fig:e19}
\end{figure*}

\label{sec:E5rev}On the stress test the true coefficients are random,
so the \claude{} prior is misaligned by construction.  At $n{=}20$ the
fixed prior is $0.095$ worse than weak in log-loss, averaged over ten
seeds (Table~\ref{tab:t1stress}, Appendix~\ref{app:addl}).  Calibration cuts this harm about
four-fold, to $0.023$; it stays significant up to $n{=}50$ and is gone
at $n{=}100$ (Table~\ref{tab:R4}g).  The robust mixture halves the remaining harm and keeps
$0.88$ and $0.92$ of the gain on Diabetes and NYC
(Figure~\ref{fig:e5rev}, Table~\ref{tab:R14}).  Two caveats.  Weak is
a near-oracle comparator here: the truth is drawn from
$\mathcal N(\bm0,0.36\bm I)$.  Against a wider one, with variance
$\tau_0^2{=}4$, only the fixed prior harms
(Table~\ref{tab:t1stress}), so the four-fold figure holds for
$\tau_0^2{=}1$ only.  And \textbf{no LLM-using method is strictly
harmless against $\mathcal N(\bm0,\bm I)$ at $n{=}20$}.

Averaging up to $20$ real draws leaves the center's displacement
unchanged (Figure~\ref{fig:e19}b), which is impossible under
independent noise.  Coarse outputs explain part of this, but the main
cause is shared \emph{errors} $\bm e_k=\bm m_k-\bm\beta^\star$.  Mean
pairwise error correlations over all draws are $0.93$--$0.99$
($0.89$--$0.99$ between the two vendors),
$0.92$--$0.99$ across paraphrases, and $0.96$ between \claude{} and a
newer flash model, \geminew, that is more accurate on California
(Tables~\ref{tab:t2para}, \ref{tab:R3}, \ref{tab:R12},
\ref{tab:R17}).  Averaging the $60$ draws from $20$ model-written
paraphrases, the AutoElicit recipe, does not repair the center either.
So log-pooling the \claude{} draws beats moment-matching on Diabetes,
where the prior is already aligned, but loses on semi-synthetic
California from $n{=}35$ and badly on the stress test (California:
Figure~\ref{fig:e8}).  The mechanism is the estimand: on collinear
designs the elicited means follow \emph{marginal} associations, not
the \emph{partial} effects they are scored against.  This shared
direction carries $97\%$ of the squared displacement on California,
and removing it drops cross-model error correlations from
$0.96$--$0.99$ to $0.09$--$0.78$ (Tables~\ref{tab:R18},
\ref{tab:R27}).  On the randomized stress test they stay high, at
$0.96$--$0.99$, a shared bias that neither prompting nor algebra
repairs.

\subsection{Post-Cutoff Benchmark, Permuted Probe}\label{sec:E18}

To exclude row-level memorization we assembled a benchmark no
eliciting model can have seen: $n{=}1{,}434$ arm's-length
one-to-three-family sales recorded in New York City between 20~May
and 30~June 2026, with $p{=}8$ hedonic features and log sale price
as the target.  The elicited prior, from \gemini{} with $K{=}3$ under
the same protocol, has the smallest displacement of the six benchmarks,
$\delta=0.455$ against $\|\bm\beta^\star\|=0.679$, and helps on data
it cannot have memorized (Table~\ref{tab:rmse};
Figure~\ref{fig:e18}): at $n{=}20$ its RMSE is $1.011$ against
$1.074$ for EB-iso, a $5.8\%$ reduction, and $1.125$ for
weak; its coefficient error is $25\%$ lower than EB-iso's.
Against the best classical method the RMSE advantage shrinks to
$3.4\%$ at $n{=}35$ and $2.2\%$ at $n{=}50$ and is gone by
$n{=}100$, while trimmed RMSE ties EB-iso even at $n{=}20$, so the
gain is in coefficient recovery and at high-leverage points.  The
anonymized prior is not weak here, so the decisive quantity is the
gap between the semantic and the anonymized prior, $5.7\%$ at
$n{=}20$; moment-matching again beats log-pooling.  Row-level
contamination is excluded; recall of stable historical relationships
is not.

To see what the prior follows, a probe pre-specified in released
code re-elicits on the same rows with neutral column
names and the true descriptions on the wrong columns.  For each
derangement $r$, $D_r=r_{\mathrm{perm}}-r_{\mathrm{id}}$ is the
correlation of the new center with the original one matched by
description, minus the same correlation matched by column.  If the
model ignored the descriptions, $D_r$ would be symmetric about zero.
A powered second stage of $20$ fresh derangements plus a
partial-permutation arm rejects this (Table~\ref{tab:R21}): $D_r>0$
in all $20$ derangements, with sign-test $p=2^{-20}$, and
column-tracking decays as more descriptions are permuted, with
Spearman $\rho=-0.55$ and $p=0.002$.  A permuted prior also harms as
predicted: displacement rises from $0.45$ to $1.09$, RMSE is $10\%$
above weak, and the gate below trusts it in at most $2.5\%$ of splits
(Table~\ref{tab:R7}).
The model follows the descriptions, not the columns.  This is
description-dependence, not reasoning, on one domain with one
elicitor.

\emph{What carries the semantics.}  The same test with the column
names kept and only the one-line descriptions shuffled, on all $28$
problems (Table~\ref{tab:t2perm}), shows that names, not
descriptions, anchor the elicitation.  On $17$ of $28$ problems the
permuted prior is the unpermuted one, with median correlation at
least $0.90$, so no misalignment is induced.  Where the names carry
no meaning, as for X1 to X8 in UCI Energy Efficiency, the prior
follows the shuffled descriptions and is misaligned: at worst the
fixed prior is $53\%$ worse than weak, the calibrated prior $8.6\%$
and the mixture $7.9\%$, and the gate's trust rate falls from $0.85$
to $0.12$.  Even when its center keeps its direction, a single-draw
fixed prior can harm by up to $80\%$, because magnitudes and scales
vary from draw to draw.

\emph{Derivation without memorization.}  Three problems generated
here separate reasoning from recall: pipe pressure drop under the
Blasius correlation, Stokes settling and a Cobb--Douglas cost
function.  The prompt states the setting and the sampling ranges but
never the formula, and the coefficients exist in no corpus
(Table~\ref{tab:t2derivable}, elicitor \gemini).  The elicited center
correlates $0.93$--$0.99$ with the coefficients the formulas imply.
Every active coefficient has the right sign and every nuisance
variable an exactly zero mean.  But the magnitudes are off, with
$\delta$ between $0.37$ and $0.58$ against the analytic vector.  At
$n{=}20$ the prior beats EB-pc on two problems, by $5.4\%$ and
$7.2\%$, and ties on the third; the advantage is gone by $n{=}50$.
Across all three tests the model contributes reasoning about signs
and zeros, not magnitudes, and not only on housing.

\subsection{Factorial, Gate, External Validation, Coverage}\label{sec:factorial}

\emph{The center carries the information, and where.}
Tables~\ref{tab:rmse} and~\ref{tab:factorial} cross the prior's
center, zero or $\mllm$, with three scales: fixed, calibrated, and
calibrated per coordinate.  By paired-bootstrap CIs, the LLM center
is significantly better than zero under the fixed scale on
semi-synthetic California, Diabetes and NYC, under the calibrated
scale on California and NYC, and under the per-coordinate scale only
on NYC.  On these three benchmarks at $n{=}20$ the LLM's own elicited
scale beats the calibrated one by $0.5$--$2.5\%$; calibration trades that for the
four-fold cut in misalignment harm of Section~\ref{sec:E5rev}.
Around the LLM center, per-coordinate scales never help significantly, and on
Diabetes they are $5.2\%$ $[3.8,6.6]$ worse than EB-iso.  On real
California the center adds no significant mean-RMSE value under any
scale.

\emph{A candidate gate.}  A parameter-free rule comes from the
matched-scale condition of Proposition~\ref{prop:adaptive}: trust the
elicited prior and moment-match if
$\|\mllm-\widehat{\bm\beta}_{\mathrm{weak}}\|\le
\|\widehat{\bm\beta}_{\mathrm{weak}}\|$, otherwise use weak.
In-suite it distrusts the misaligned stress prior in $91\%$ of splits
and trusts the aligned benchmarks in $45$--$100\%$
(Table~\ref{tab:R9}).  Under sign surgery its trust
rate falls from $84\%$ to $6\%$ (Table~\ref{tab:R16}).  We chose its
form after seeing all six benchmarks and froze it, with the selection
rule, before retrieving any external dataset.  The external protocol
below is its only test.

\emph{Pre-specified external validation.}  The rule of
Appendix~\ref{app:t2}, written before any candidate was
retrieved, selected six UCI regression problems without human choice
(Table~\ref{tab:t2datasets}; elicitor \gemini).  The elicited centers
are again coarse, with $\delta$ from $0.38$ to $1.05$ and sign
agreement from $0.50$ to $0.88$.  Against the strongest zero-centered
classical prior in each cell (Table~\ref{tab:t2frozen}), the
moment-matched prior at $n{=}20$ is better on four problems: Concrete
by $4.7\%$, Energy Efficiency $7.4\%$, Istanbul Stock Exchange $1.7\%$
and Bike Sharing $2.8\%$.  It is $0.6\%$ worse on Forest Fires and
$17.5\%$ worse on Air Quality.  The Air Quality target, absolute
humidity, is nearly determined by two features, so
$\|\bm\beta^\star\|$ is large, the gate's relative threshold is
lenient, and the gate trusted every split.  The pre-stated verdict is
\textsc{partial}: gated relative regret is $2.4\%$ on distrusted and
$3.2\%$ on trusted cells, within the $3\%/9\%$ envelopes, but one
dataset violates the no-trusted-harm criterion.  The robust mixture
needs no gate.  Its relative regret is at most $1.4\%$ on every cell
except Energy at $n{=}20$, where it is $6.5\%$ because the moment
prior is itself the oracle, and it averages $1.0\%$ against $3.0\%$
for the gated policy.  The calibrated prior cuts the Air Quality harm
to $+4.3\%$, again four-fold.  At $n{=}50$ the moment-matched
prior's gains shrink to $1$--$3\%$ and its Air Quality harm to
$+5.3\%$, and Forest Fires is $2.3\%$ worse.

\emph{Pre-specified extension.}  Fourteen further rule-selected
datasets from OpenML-CTR23 and PMLB (Table~\ref{tab:t2extension})
bring the total to $18$ distinct external problems, with two OpenML
copies of frozen datasets counted once by their frozen-run results.
At $n{=}20$ the center beats
the strongest classical prior with a CI excluding zero on six, by
$1.7$ to $7.4\%$: Concrete, Energy, Istanbul, Bike Sharing, Auction
Verification and BNG-pharynx.  On eight it is within $-1.2$ to
$+3.7\%$.  On four it is worse by more than $5\%$, from $8.8$ to
$57\%$: Air Quality, space\_ga, BNG-echoMonths and pollen.  The last
two are synthetic data whose descriptions do not describe the
generating process.  The verdict is again \textsc{partial}, with
space\_ga the one violating dataset.  The gate distrusts pollen and
echoMonths on every split but trusts space\_ga and Air Quality.  The
mixture's relative regret stays at or below $7.5\%$ on all $40$
external cells.

\emph{Coverage.}  At $n{=}20$ nominal-$90\%$ predictive intervals
under-cover on average for every method, from $0.889$ for weak to
$0.825$ for the log pool (Table~\ref{tab:R13}).  Coefficient
intervals differ far more: flat and weak cover $0.89$--$0.94$ of the
coefficients, EB-iso $0.59$--$0.83$, LLM-fixed $0.60$--$0.81$,
LLM-cal $0.33$--$0.71$ and the per-coordinate priors at most $0.38$
(Table~\ref{tab:t1coefcov}).

\section{Discussion}\label{sec:disc}

Four practical rules follow.

\textbf{(1)~Run the anonymization control and read it as
behavior.}  Without names both models fall back to zero means, so the
gain comes from the names.  The control cannot tell understanding
from recall.

\textbf{(2)~Do not read the elicited scale as uncertainty.}  The
elicited $95\%$ intervals cover between $0.10$ and $1.00$ of the true
coefficients (Table~\ref{tab:R26}).  As shrinkage, at $n{=}20$ the
elicited scale beat the calibrated one on the three aligned in-suite
benchmarks, but was four times more harmful on the misaligned stress
test, at $\tau_0^2{=}1$.  Unless you
can check the center's displacement, use the calibrated scale or the
robust mixture, which kept relative regret within $7.5\%$ on all $40$
external cells against $13.9\%$ for the hard gate.

\textbf{(3)~Moment-match, and treat every draw, paraphrase and vendor
as dependent.}  Errors correlate $0.92$--$0.99$ across all three, so
summing precisions counts the same evidence several times.

\textbf{(4)~Report against the strongest adaptive classical prior,
under more than one loss, and state the comparator's scale.}  On
heavy-tailed data the losses rank the methods differently
(Table~\ref{tab:losses}), and the same prior can look harmful or
harmless depending on the comparator's variance $\tau_0^2$
(Table~\ref{tab:t1stress}).

\section{Conclusion}

The negative results are firm.  Errors are shared across draws,
paraphrases and vendors, elicited scales are not calibrated, and
treating draws as independent evidence creates false certainty.  The
positive result is narrow.  An elicited center supplies roughly
population-level marginal knowledge without any data, not partial
effects.  It helps clearly on one real problem and marginally on
another, but not where per-coordinate structure or extreme leverage
dominates, and the ranking depends on the loss.  The pre-specified
external test returned \textsc{partial}: on $18$ rule-selected
problems the center helps on six and hurts on four.  The hard gate is
not a safe guard: its relative regret reaches $13.9\%$, while the
evidence-weighted mixture stays within $7.5\%$ on all $40$ external
cells.  Elicitation is worth its $K$ API calls only when an
interpretable posterior over coefficients is needed and the center is
plausibly close to the truth.  Otherwise zero-centered adaptive
shrinkage predicts as well or better.  A prediction-only TabPFN reference \citep{hollmann2025} trails the
best prior-based method in mean RMSE by $5$ to $10\%$ on three of
four benchmarks at $n{=}20$ and leads on real California, $0.817$
against $1.726$ for EB-pc, but only in mean RMSE
(Tables~\ref{tab:rmse}--\ref{tab:losses}).
\label{endofcontent}

\section*{Limitations}\label{sec:limits}

\textbf{Model coverage.}  Only the five in-suite benchmarks other than
NYC-2026 used the frontier model, \claude{} at $K{=}3$.  All other experiments, and so
every LLM-specific conclusion, rest on one flash-tier model with
coarse outputs, \gemini, and one prompt style.  No model was added
after the review, so error correlations between frontier models are
untested; their persistence, which marginal-effect reporting
suggests, is an expectation, not a result.  More frontier draws would
hardly move the center: displacement is flat from $K{=}1$ to $20$
(Figure~\ref{fig:e19}b).
\textbf{Scope.}  Results cover public, documented tabular regression
with named features, $4\le p\le10$ and $n\le150$, drawn by a stated
rule from UCI, OpenML-CTR23 and PMLB.  Harm concentrated on synthetic
or near-deterministic targets.  Larger $p$ and non-tabular settings are
untested, and without feature names the elicited prior is
uninformative (Section~\ref{sec:E4}).
\textbf{Value-add, breadth, loss.}  The gain over the best
zero-centered prior is mixed: positive on $4$ of $6$ in-suite
problems (sign test $p=0.69$; Table~\ref{tab:t1cross}), with $6$
helped, $8$ within $\pm4\%$ and $4$ harmed externally, and
loss-dependent on heavy-tailed targets.
\textbf{Comparator dependence of harm.}  Harm relative to weak
depends on the weak prior's scale (Section~\ref{sec:E5rev}).
Real-data harm occurs at $n{=}20$: $16\%$ higher log-loss for
\gemini{} on Breast Cancer.
\textbf{Semantics vs.\ memorization.}  Anonymization rules out only
generic shrinkage and NYC-2026 only row-level contamination.  The
probes show dependence on names and descriptions; the derivable
problems show reasoning about signs and zeros, not magnitudes.
\textbf{Theory scope.}  All results are classical identities.  The
logistic extension is approximate, and the only guarantee is
dominance over the flat prior in the orthogonal design
(Section~\ref{sec:theory}).
\textbf{Pre-specification.}  The third-party registration announced
in the submitted version was not done.  Instead, the selection rule
was frozen before any candidate was retrieved and its addendum
before any Tier-2 call; both files ship with the
code (Appendix~\ref{app:t2}).
\textbf{Pseudo-truth and estimand.}  On real data $\bm\beta^\star$ is
the full-data OLS fit (the MAP fit for Breast Cancer), so $\delta$ is uncertain, $0.76\pm0.26$ on
Diabetes, and on collinear designs part of it is estimand mismatch
(Section~\ref{sec:E4}).

\section*{Ethical Considerations}
Elicited prior centers carry the associations present in the
eliciting model's training corpus.  In the domains used here, housing
prices, health and finance-like targets, such associations can
reflect historical discrimination or proxies for protected
attributes.  Nothing in the audit detects this: anonymization removes
all semantics, and the trust gate and the robust mixture only compare
the elicited center with the weak posterior on the data at hand, so a
prior that is biased in the same direction as the data is trusted
without being unbiased.  Elicited priors should therefore not be used
for coefficients whose sign or size carries fairness implications
without a domain audit of the rationale strings the protocol
records.  All datasets are public and contain no personal
identifiers: the NYC-2026 benchmark uses hedonic features
and prices from public rolling-sales records, and all datasets are
used under the terms of their sources, which the code repository lists.  No human subjects were involved,
and API usage was limited to the recorded elicitations.

\bibliography{references}

\appendix
\renewcommand{\topfraction}{0.92}
\renewcommand{\bottomfraction}{0.85}
\renewcommand{\textfraction}{0.05}
\renewcommand{\floatpagefraction}{0.85}
\renewcommand{\dbltopfraction}{0.92}
\renewcommand{\dblfloatpagefraction}{0.85}
\setcounter{topnumber}{5}\setcounter{bottomnumber}{3}
\setcounter{totalnumber}{8}\setcounter{dbltopnumber}{5}
\setlength{\textfloatsep}{12pt plus 2pt minus 3pt}\setlength{\dbltextfloatsep}{12pt plus 2pt minus 3pt}
\setlength{\floatsep}{9pt plus 2pt minus 2pt}\setlength{\dblfloatsep}{9pt plus 2pt minus 2pt}

\section{Elicitation Prompt}\label{app:prompt}
The template is filled per dataset; the wording below is verbatim,
with the ruling lines of the layout omitted.  In the anonymized variant
the feature names are \texttt{feature\_j}, every feature is described
as ``a standardized real-valued feature'', and the dataset description
is one content-free sentence.
Learned features such as graph embeddings \citep{zou2023graph} have no
names to begin with.

\begin{quote}\small\ttfamily\raggedright
SYSTEM: You are an expert statistician.  You answer with numerical
hyperparameters for Gaussian priors and a one-sentence rationale per
coefficient.  You do NOT see the dataset.\par
USER: I am performing a Bayesian linear regression that predicts a
standardized response (approximately N(0,1) marginally) from the
following standardized features (each z-scored across the population
of interest).  For each coefficient beta\_j please return an
informative Gaussian prior N(mu\_j, sigma\_j\textasciicircum 2)
reflecting your best guess of sign and magnitude.  After returning the
priors, also return a single JSON object on its own line of the form
\{"feature": [mu, sigma], ...\}\par
Dataset description: $\langle$dataset description$\rangle$\par
Features: $\langle$one line per feature: name and description$\rangle$\par
Please return one sentence of reasoning per feature, then the JSON.
\end{quote}

The template, its per-dataset instantiations and the recorded
responses are in \texttt{code/priors.py} and
\texttt{code/gap\_elicitations.json}; the parsed
trailing JSON is what enters the experiments, so the elicitation step
is auditable and reproducible without any model access.
The one-sentence rationales are stored but not scored; explanations can
carry an audit signal of their own \citep{liu2026explanations}.

\section{Proof of Theorem~\ref{thm:bv}}\label{app:thm1}

\begin{proof}
From $\bm S_n^{-1}=\bm S_0^{-1}+\sigma^{-2}\bm X^\top\bm X$ and
$\E[\bm X^\top\bm y\mid\bm X]=\bm X^\top\bm X\bm\beta^\star$,
$\E[\bm m_n]=\bm S_n(\bm S_n^{-1}\bm\beta^\star+\bm S_0^{-1}\bm\delta)
=\bm\beta^\star+\bm S_n\bm S_0^{-1}\bm\delta$.  Since $\bm m_n$ is
linear in $\bm y$ with coefficient $\sigma^{-2}\bm S_n\bm X^\top$,
$\Cov(\bm m_n)=\sigma^{-2}\bm S_n\bm X^\top\bm X\bm S_n
=\bm S_n(\bm S_n^{-1}-\bm S_0^{-1})\bm S_n$.  Then
$\MSE=\lVert\bias\rVert^2+\tr\Cov$.
\end{proof}

\section{Proof of Corollary~\ref{cor:cross}}\label{app:cor}
With $\bm X^\top\bm X=n\bm I$ and $\bm S_0=\tau^2\bm I$,
$\bm S_n=\frac{\tau^2\sigma^2}{\sigma^2+n\tau^2}\bm I$; the bias on
coordinate $j$ from Theorem~\ref{thm:bv} is
$\frac{\sigma^2}{\sigma^2+n\tau^2}\delta_j$ and the variance is
$\frac{n\tau^4\sigma^2}{(\sigma^2+n\tau^2)^2}$,
giving \eqref{eq:msej}.  The flat limit has MSE $\sigma^2/n$.
Multiplying $\MSE_j(n)\le\sigma^2/n$ through by
$n(\sigma^2+n\tau^2)^2>0$ yields
$n\sigma^4\delta_j^2+n^2\sigma^2\tau^4\le\sigma^6+2n\sigma^4\tau^2+n^2\sigma^2\tau^4$,
i.e.\ $n(\delta_j^2-2\tau^2)\le\sigma^2$.  If
$\delta_j^2\le2\tau^2$ this holds for all $n$; otherwise the
threshold is $n^\star_j=\sigma^2/(\delta_j^2-2\tau^2)$. \qed

\section{Proof of Theorem~\ref{thm:cal}}\label{app:thm2}
Let $\bm A:=\sigma^2\bm I+(1/\alpha)\bm X\bm X^\top$ and
$\bm r:=\bm y-\bm X\mllm$.  Then
$\partial_\alpha\log p
=-\tfrac12[\tr(\bm A^{-1}\partial_\alpha\bm A)
-\bm r^\top\bm A^{-1}(\partial_\alpha\bm A)\bm A^{-1}\bm r]$
with $\partial_\alpha\bm A=-(1/\alpha^2)\bm X\bm X^\top$.  By the
Woodbury identity, $\bm A^{-1}\bm X=(\alpha/\sigma^2)\bm X\bm B^{-1}$
with $\bm B:=\bm X^\top\bm X/\sigma^2+\alpha\bm I$, whence
$\tr(\bm A^{-1}\bm X\bm X^\top)=\alpha\gamma(\alpha)$, and
$\bm m_{\mathrm{MAP}}-\mllm=\bm B^{-1}\bm X^\top\bm r/\sigma^2$ gives
$\bm r^\top\bm A^{-1}\bm X\bm X^\top\bm A^{-1}\bm r
=\alpha^2\lVert\bm m_{\mathrm{MAP}}-\mllm\rVert^2$.  Substituting,
\[
\partial_\alpha\log p
=\frac1{2\alpha}\big[\gamma(\alpha)
-\alpha\lVert\bm m_{\mathrm{MAP}}(\alpha)-\mllm\rVert^2\big],
\]
and stationarity gives \eqref{eq:alphastar}.

\emph{Monotone EM.}  View $\bm\beta$ as the latent variable of the
model $\bm\beta\sim\mathcal N(\mllm,\alpha^{-1}\bm I)$,
$\bm y\mid\bm\beta\sim\mathcal N(\bm X\bm\beta,\sigma^2\bm I)$, with
parameter $\alpha$.  The E-step posterior is
$\mathcal N(\bm m_{\mathrm{MAP}}(\alpha^{(t)}),\bm S_n(\alpha^{(t)}))$;
the M-step maximizer of the expected complete-data log-likelihood is
$\alpha^{(t+1)}
=p/(\lVert\bm m_{\mathrm{MAP}}(\alpha^{(t)})-\mllm\rVert^2+\tr\bm S_n(\alpha^{(t)}))$.
By \citet{dempster1977} each step weakly increases the marginal
likelihood, and by the MacKay identity
$\gamma(\alpha)=p-\alpha\tr\bm S_n(\alpha)$ the EM and MacKay updates
share fixed points.

\emph{Non-concavity and uniqueness.}  In the eigenbasis of
$\bm X^\top\bm X$ with eigenvalues $\lambda_j$ and residual
coordinates $r_j$,
\begin{align*}
\log p(\alpha)=C-\tfrac12\sum_j\Big[&\log(\sigma^2+\lambda_j/\alpha)\\
&+\frac{r_j^2}{\sigma^2+\lambda_j/\alpha}\Big].
\end{align*}
The first term in the bracket contributes a concave function of
$\log\alpha$; the second
contributes
curvature whose sign flips with $\lambda_j/\alpha\gtrless\sigma^2$,
so global concavity fails in general (verified numerically in the
released unit tests).  A sufficient condition for at most one stationary
point is that $\bm X^\top\bm r$ lies in the eigenspace of the largest
eigenvalue of $\bm X^\top\bm X$; the general characterization, in the spirit of
\citet{wipf2008}, is open.  Empirically the marginal likelihood is single-peaked
on all our designs. \qed

\section{Adaptive Calibration in the Orthogonal Design}\label{app:adaptive}

\begin{proposition}[Adaptive calibration, orthogonal design]
\label{prop:adaptive}
Let $\bm X^\top\bm X=n\bm I$ and $\bm z=\bm X^\top\bm y/n$, so
$\bm z\sim\mathcal N(\bm\beta^\star,(\sigma^2/n)\bm I)$.  The
type-II ML problem has the closed-form solution
$\widehat{\tau}^{\,2}=\bigl(\lVert\bm z-\mllm\rVert^2/p-\sigma^2/n\bigr)_+$,
and the resulting posterior mean is positive-part linear shrinkage
of the OLS toward the elicited center,
\[
\widehat{\bm\beta}
=\mllm+\Bigl(1-\tfrac{p\,\sigma^2/n}{\lVert\bm z-\mllm\rVert^2}\Bigr)_{\!+}
(\bm z-\mllm),
\]
a translated Efron--Morris/James--Stein-type estimator.
Consequently, for $p\ge5$ it dominates the flat/OLS estimator in
total squared error uniformly in $\bm\beta^\star$
\citep{james1961,efron1973}.  No analogous guarantee holds relative
to the weak prior or empirical Bayes: with matched scales
($\tau=\tau_0$) the fixed LLM-centered prior beats the weak prior
at every $n$ on coordinate $j$ exactly when
$\delta_j^2\le\beta_j^{\star2}$, because both estimators shrink
identically and only the squared center displacement differs; the
adaptive version can and does underperform weak under
misalignment.  What remains open is a finite-sample uniform
oracle inequality for correlated designs,
$R(\widehat{\bm\beta}_{\widehat\alpha})\le
\inf_{\alpha\ge0}R(\widehat{\bm\beta}_\alpha)
+\mathrm{cost}(n,p,\sigma^2)$, presumably via a SURE- or
concentration-based analysis in the tradition of \citet{stein1981}
and \citet{xie2012}.
\end{proposition}

\section{Proof of Theorem~\ref{thm:logpool}}\label{app:thm3}
\noindent\emph{Formal statement.}  If $\bm m_k=\bm\beta+\bm\varepsilon_k$ with
$\bm\varepsilon_k\sim\mathcal N(\bm0,\bm S_k)$ independent across $k$
and a flat hyper-prior on $\bm\beta$, the posterior of $\bm\beta$
given $\{\bm m_k\}$ is exactly \eqref{eq:lp}; for equal
$\bm S_k=\bm S$ the pooled covariance is $\bm S/K$, strictly
below $\bm S_{\mathrm{mm}}$ in the PSD order for $K\ge2$.  If instead
$\bm\varepsilon_k=\bm b+\bm\eta_k$ with a fixed shared bias
$\bm b\neq\bm0$, the pooled mean converges to $\bm\beta+\bm b$ while
the pooled covariance still shrinks as $1/K$: the pool becomes
confidently wrong, in that every fixed-level credible ellipsoid of the
pool covers $\bm\beta$ with probability tending to zero.  (Proof below.)

\noindent\emph{Proof.}
Under $\bm m_k=\bm\beta+\bm\varepsilon_k$,
$\bm\varepsilon_k\sim\mathcal N(\bm0,\bm S_k)$ independent, with flat
hyper-prior,
$p(\bm\beta\mid\{\bm m_k\})\propto
\prod_k\exp(-\tfrac12(\bm m_k-\bm\beta)^\top\bm S_k^{-1}(\bm m_k-\bm\beta))$;
completing the square gives precision $\sum_k\bm S_k^{-1}$ and mean
$(\sum_k\bm S_k^{-1})^{-1}\sum_k\bm S_k^{-1}\bm m_k$, i.e.\
\eqref{eq:lp}.  For $\bm S_k\equiv\bm S$,
$\bm S_{\mathrm{lp}}=\bm S/K\preceq\bm S\preceq\bm S_{\mathrm{mm}}$,
strictly for $K\ge2$.  Under the common-bias model
$\bm\varepsilon_k=\bm b+\bm\eta_k$ with
$\bm\eta_k\sim\mathcal N(\bm0,\bm S)$ i.i.d., the pooled mean is
$\bm\beta+\bm b+\bar{\bm\eta}$ with
$\bar{\bm\eta}\sim\mathcal N(\bm0,\bm S/K)$, so
$\E\lVert\bm m_{\mathrm{lp}}-\bm\beta\rVert^2
=\lVert\bm b\rVert^2+\tr(\bm S)/K\to\lVert\bm b\rVert^2$, while the
pooled covariance $\bm S/K\to\bm0$: the aggregate concentrates on the
wrong value.  The level-$c$ pooled ellipsoid contains $\bm\beta$ iff
$K(\bm b+\bar{\bm\eta})^\top\bm S^{-1}(\bm b+\bar{\bm\eta})\le c$;
since $\bar{\bm\eta}\to\bm0$ almost surely and $\bm b\neq\bm0$, the
left side diverges and the coverage tends to zero.  The risk of the
pool as a prior is not monotone in $K$: under the conditions of
Corollary~\ref{cor:cross} with $\bm S=s\bm I$, \eqref{eq:msej} at
$\tau^2=s/K$ has
$\partial_{\tau^2}\MSE_j=2n\sigma^4(\tau^2-\delta_j^2)/(\sigma^2+n\tau^2)^3$,
so it falls while $s/K>\delta_j^2$ and rises afterwards. \qed

\section{Extension to Logistic Regression: Approximate Fixed Point}\label{app:thm4}

For binary $y_i$ with
$p_i(\bm\beta)=\sigma(\bm x_i^\top\bm\beta)$ and prior
$\bm\beta\sim\mathcal N(\mllm,\alpha^{-1}\bm I)$, let
$\bm\beta_{\mathrm{MAP}}(\alpha)$ be the MAP and
$\bm H(\alpha)=\bm X^\top\bm W_{\mathrm{MAP}}\bm X+\alpha\bm I$ the
negative log-posterior Hessian,
$\bm W_{\mathrm{MAP}}=\mathrm{diag}(p_i(1-p_i))$.  The Laplace log
evidence replaces \eqref{eq:marglik}.

\begin{proposition}[Approximate (Laplace/MacKay) fixed point for the GLM]\label{thm:logcal}
Any stationary point of the Laplace evidence satisfies
\begin{align}
\alpha^\star&=\gamma_{\mathrm{lap}}(\alpha^\star)\big/
\lVert\bm\beta_{\mathrm{MAP}}(\alpha^\star)-\mllm\rVert^2,\label{eq:loggam}\\
\gamma_{\mathrm{lap}}(\alpha)&=p-\alpha\tr\bm H(\alpha)^{-1},\nonumber
\end{align}
under the MacKay envelope approximation that treats the indirect
dependence of $\bm W_{\mathrm{MAP}}$ on $\alpha$ as higher-order.  The
resulting fixed-point iteration (inner MAP refit, outer $\alpha$
update) ascends a Laplace surrogate; we do not claim exact monotone
ascent of the true evidence.  Empirically it stops within its cap of $200$
iterations in most splits.  (Derivation below; the statement is an approximation, not a theorem.)
\end{proposition}

\noindent\emph{Derivation.}
Write the Laplace evidence
$\mathcal E(\alpha)=\ell(\bm\beta_{\mathrm{MAP}})
-\tfrac\alpha2\lVert\bm\beta_{\mathrm{MAP}}-\mllm\rVert^2
+\tfrac p2\log\alpha-\tfrac12\log|\bm H(\alpha)|+\text{const}$, where
$\ell$ is the Bernoulli log-likelihood.  Differentiating in $\alpha$,
the indirect term through $\bm\beta_{\mathrm{MAP}}(\alpha)$ vanishes
by MAP stationarity (envelope theorem); treating
$\partial_\alpha\bm W_{\mathrm{MAP}}$ as higher-order (MacKay's
approximation) gives
$\partial_\alpha\log|\bm H|\approx\tr(\bm H^{-1})$ and
\[
\partial_\alpha\mathcal E\approx\tfrac p{2\alpha}
-\tfrac12\tr(\bm H^{-1})
-\tfrac12\lVert\bm\beta_{\mathrm{MAP}}(\alpha)-\mllm\rVert^2,
\]
whose zero is \eqref{eq:loggam}.  The inner MAP refit is a partial
M-step and the outer update the closed-form maximizer of the
surrogate; we observe convergence in most splits and claim no
monotone ascent of the true evidence. \qed

\section{Logistic Calibration Check (E9)}\label{app:theoryexp}

\begin{figure*}[tp]\centering
\includegraphics[width=0.92\textwidth]{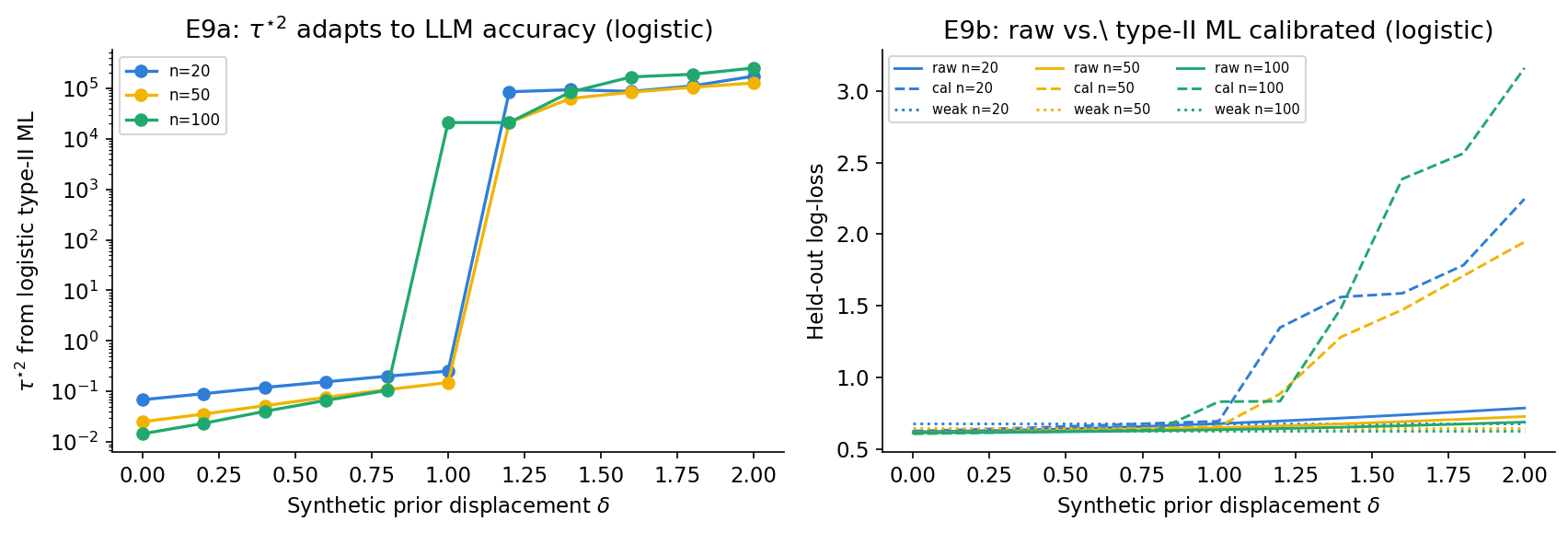}
\caption{E9 (synthetic displacement sweep, no LLM): the calibrated scale of the logistic (Laplace) version tracks displacement; it improves on the fixed scale only at small displacement and $n\ge50$.}
\label{fig:e9}
\end{figure*}

In a synthetic displacement sweep without any LLM
(Figure~\ref{fig:e9}), the logistic calibration of
Proposition~\ref{thm:logcal} adapts monotonically from median $\tau^{\star2}\approx0$
at $\delta{=}0$ to $0.09$ at $\delta{=}0.8$ and $0.21$ at
$\delta{=}1.2$ at $n{=}20$, the correct qualitative response to
misalignment, although at $n{=}20$ it does not improve on the fixed scale.

\section{Cross-Vendor Replication, Prompt Sensitivity and Hierarchical Pooling (E16, E17, E19)}\label{app:crossmodel}

\paragraph{A second vendor (E16).}  We repeated the elicitation with
\gemini{} ($K{=}3$ per prompt, same templates).  Its means are close
to \claude{}'s (Table~\ref{tab:R17}): displacements of $1.099$ vs.\
$1.095$ on California, $0.760$ vs.\ $0.760$ on Diabetes, $0.987$ vs.\
$1.066$ on the stress test and $3.112$ vs.\ $2.696$ on Breast Cancer,
with the same sign agreement on three of the four.  The
between-vendor correlation of elicited means is $0.985$ on California
and $0.991$ on the stress test, as large as the within-model
correlations ($0.983$, $0.992$), so cross-vendor draws are not
independent evidence there.  At $n{=}20$ the \gemini{} prior gives
RMSE $0.684$ on California and $0.822$ on Diabetes, against $0.691$
and $0.846$ for \claude.  On Breast Cancer the vendors diverge: the
\gemini{} prior hurts (log-loss $0.323$ vs.\ weak
$0.279$ at $n{=}20$) while \claude{}'s helps ($0.247$).  Displacement,
not vendor identity, decides the sign of the effect, as
Theorem~\ref{thm:bv} prescribes.

\paragraph{Prompt and temperature sensitivity (E17).}  With five fixed
paraphrases of the prompt and temperatures $0.2$, $0.7$ and $1.0$
(\gemini, three draws per cell), the displacement stays within
$[1.099,1.287]$ on California and $[0.653,0.760]$ on Diabetes across
paraphrases and moves by at most $0.10$ across temperatures; sign
agreement never falls below $0.60$.  Errors across paraphrases
correlate as strongly as errors across repeated draws of one prompt
(Table~\ref{tab:R12}).

\paragraph{Hierarchical pooling (E19).}  The hierarchical pool of
Figure~\ref{fig:e19} models draw $k$ of model $m$ as
$\bm m_{mk}=\bm\beta+\bm b_m+\bm\eta_{mk}$ with a per-model bias
$\bm b_m\sim\mathcal N(\bm0,\bm\Sigma_b)$.  Under a flat hyper-prior
its posterior covariance cannot fall below $\bm\Sigma_b/M$ as
$K\to\infty$, so repeated sampling of one model never shrinks
uncertainty below the between-model spread.  This removes the
within-model over-sharpening of the log pool
(Theorem~\ref{thm:logpool}) and nothing more: a bias $\bm b_0$ common
to all vendors enters the elicitations only through
$\bm\beta+\bm b_0$, so no aggregator of elicitations alone is
consistent for $\bm\beta$.  $\bm\Sigma_b$ is estimated by one-way
ANOVA when $M\ge2$; with $M{=}1$ the estimator falls back to moment
matching.  In the simulation of Figure~\ref{fig:e19}a, the log pool's
nominal-$90\%$ coverage of $\bm\beta$ falls from $0.70$ to $0.33$ as
$K$ grows from $1$ to $10$, the oracle hierarchical pool stays at
$0.90$, the estimated version falls from $0.86$ to $0.68$ with only
$M{=}2$ groups, and moment matching over-covers at $0.99$.  With
$K{=}20$ recorded \gemini{} draws per benchmark
(Figure~\ref{fig:e19}b), the displacement of the averaged mean is
flat from $K{=}1$ to $20$ (California $1.15\to1.16$, Diabetes
$0.72\to0.72$, stress test $0.98\to0.97$), and the log pool's coverage
of $\bm\beta^\star$ collapses (Diabetes $0.76\to0.10$, California
$0.42\to0.00$) while moment matching stays flat.  The hierarchical
pool collapses with it under a within-vendor grouping (Diabetes
$0.64\to0.12$), because the shared component is untouched.
Agreement across parallel generations, an uncertainty signal in other
settings \citep{xue-etal-2026-reason}, is therefore not a sign of
accuracy here.

\section{Additional Analyses}\label{app:addl}

\paragraph{Stress family (E5).}  Unless noted, stress-test numbers use ten DGP seeds
with identical paired splits, since single-seed results can
be unstable in limited-data settings \citep{zhan2026variances}.  The gate evaluation of
Section~\ref{sec:factorial} uses the nine seeds drawn independently of
the recorded elicitation at two training sizes, which
gives $18$ cells, $17$ of them misaligned; the tenth is the DGP under
which the elicitation was recorded.  The robust mixture's residual
harm is small, $+0.010$ at $n{=}20$ and $+0.003$ at $n{=}50$
(Table~\ref{tab:t1stress}), and not significant from $n{=}35$ in the
paired tests of Table~\ref{tab:R4}g ($230$ evaluations per cell: $50$ splits
for one seed and $20$ for each of the other nine).  Its posterior weight on the
LLM component is about $0.54$: the data at these sizes cannot fully
reject the misaligned component.  Retention on the aligned side is
the mixture's gain over weak divided by the best LLM method's gain on
the shared $100$-split grid (split bootstrap, $B{=}2000$): $0.88$
$[0.83,0.92]$ on Diabetes and $0.92$ $[0.83,1.04]$ on NYC
(Table~\ref{tab:R14}).  On semi-synthetic California the gain is too
small for the ratio to be estimated ($0.26$ $[-1.28,0.63]$), so we
make no retention claim there.

\paragraph{Sign surgery.}  Flipping the signs of the $k$
largest-$|\beta^\star_j|$ coordinates of the accurate Diabetes prior
erodes its gain (fixed isotropic scale $0.3$), from RMSE $0.861$ at $k{=}0$ to $0.898$ at $k{=}4$
against $0.989$ for weak, without crossing into harm at $n{=}20$,
while the gate's trust rate falls from $84\%$ to $6\%$
(Table~\ref{tab:R16}).

\paragraph{Mechanical checks on the dependence measurement (E11).}
Elicited means are heavily discretized ($98\%$ lie on a $0.05$ grid;
only $28$ distinct values among $306$ elicitations): the models
report stylized textbook magnitudes, not continuous
estimates.  But jittering uniformly within grid cells leaves every
correlation essentially unchanged ($0.990\!\to\!0.989$ on
California), so granularity does not drive the
dependence.  Verbalized confidence scores show the same rounding
\citep{dai2026rescaling}.  Because real-data errors are scored against a common
full-data OLS pseudo-truth, we propagated pseudo-truth uncertainty
($B{=}500$ row bootstraps) and subtracted the
shared-estimation-error term in closed form: corrected correlations
move by at most $0.03$ (Diabetes $0.93\!\to\!0.91$,
$0.98\!\to\!0.97$), and the stress-test and
semi-synthetic measurements are exact by construction, the principle
behind ProbabilityBench \citep{dai2026probabilitybench}.  The
within-vendor intraclass correlation of errors beyond the common
component ranges from $0.16$ (California) to $0.91$ (logistic).
Error dependence alone does not predict whether pooling helps: the
shared component is near-zero error on Diabetes and large error on
the stress test.

\paragraph{Estimand decomposition: details (E11; Tables~\ref{tab:R18}, \ref{tab:R27}, \ref{tab:R22}).}
On real, collinear designs, elicited means correlate
$0.89$--$0.97$ with per-feature simple-regression slopes vs.\ only
$0.44$--$0.61$ with the OLS $\bm\beta^\star$ they are scored
against (Diabetes).  The shared marginal-vs-partial discrepancy
direction carries $97\%$ of squared displacement on California
(random-direction null $12.5\%$, $p<10^{-4}$; cross-fit fraction
$97\%$); on Diabetes the in-sample fraction is $76\%$ (null
$p=2\times10^{-4}$) but the estimated direction is unstable under
cross-fitting, so the estimand mechanism is claimed only for the
strongly collinear design.  Re-eliciting with the full feature
correlation matrix disclosed and an explicit ceteris-paribus
(partial-effect) instruction leaves the marginal-tracking signature
of real California essentially unchanged (fraction of $\delta^2$
along the discrepancy direction $0.98\!\to\!0.96$; correlation with
$\bm\beta^\star$ slightly \emph{worse}) and only modestly reduces
NYC displacement (Table~\ref{tab:R22}).  Both components are shared, so
neither rescues precision-summing; but the mechanism matters for
what diversification could ever buy, and we scope our attribution
accordingly.

\subsection*{Robustness Checks (E6, E12)}\label{sec:exp-robust}

\emph{Unknown $\sigma^2$ (E6).}  The full NIG analysis with
Student-$t$ predictive leaves point predictions unchanged (as it must
in the conjugate family) and gives central-$90\%$ coverage within
$0.03$ of nominal for the semantic prior at $n\ge35$; a
prior-agnostic estimate of $\sigma^2$ changes no benchmark's top
method (Table~\ref{tab:t1sigma}).
\emph{Horseshoe (E12).}  The horseshoe improves on weak where the
truth is sparse (Diabetes $\approx0.95$ vs.\ weak $0.99$ at $n{=}20$)
but does not match the LLM prior there ($0.83$--$0.85$), and EB-pc
remains the stronger classical adaptive prior; the practical
``no-LLM wins'' case remains real California (E1b).

\section{Additional Figures}\label{app:figs}

\begin{figure*}[tp]\centering
\begin{minipage}[t]{0.48\textwidth}\centering
\includegraphics[width=0.8\linewidth]{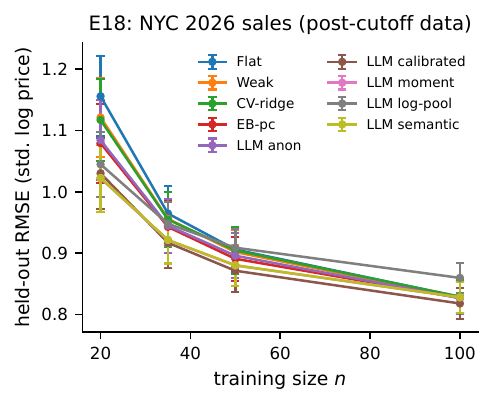}
\caption{E18 (\gemini, $K{=}3$): held-out RMSE on NYC-2026 (95\% CIs, 100 splits). Every sale post-dates the models' training data, which rules out memorized rows but not remembered relationships.}
\label{fig:e18}
\end{minipage}\hfill
\begin{minipage}[t]{0.48\textwidth}\centering
\includegraphics[width=0.95\linewidth]{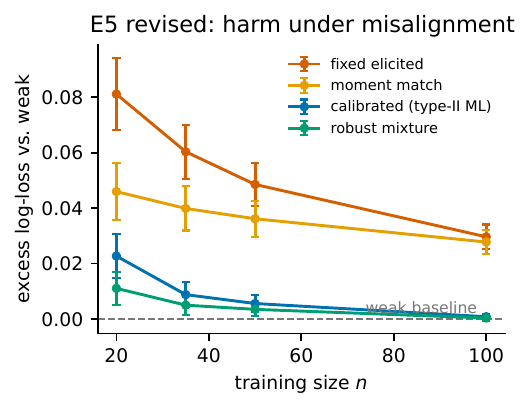}
\caption{E5 (\claude{} prior, $K{=}3$): harm on the misaligned stress test, measured as held-out log-loss above the weak prior $\mathcal N(\bm0,\bm I)$ (ten DGP seeds, paired splits, 95\% CIs; positive $=$ harm). The moment-match curve pools both vendors' draws ($K{=}6$). Calibration cuts the fixed prior's harm about four-fold and the robust mixture halves it again, but none reaches the weak baseline at $n{=}20$ (50-split re-run: $+0.095$/$+0.023$/$+0.010$, Table~\ref{tab:t1stress}).}
\label{fig:e5rev}
\end{minipage}
\end{figure*}

\begin{figure*}[tp]\centering
\begin{minipage}[t]{0.48\textwidth}\centering
\includegraphics[width=0.8\linewidth]{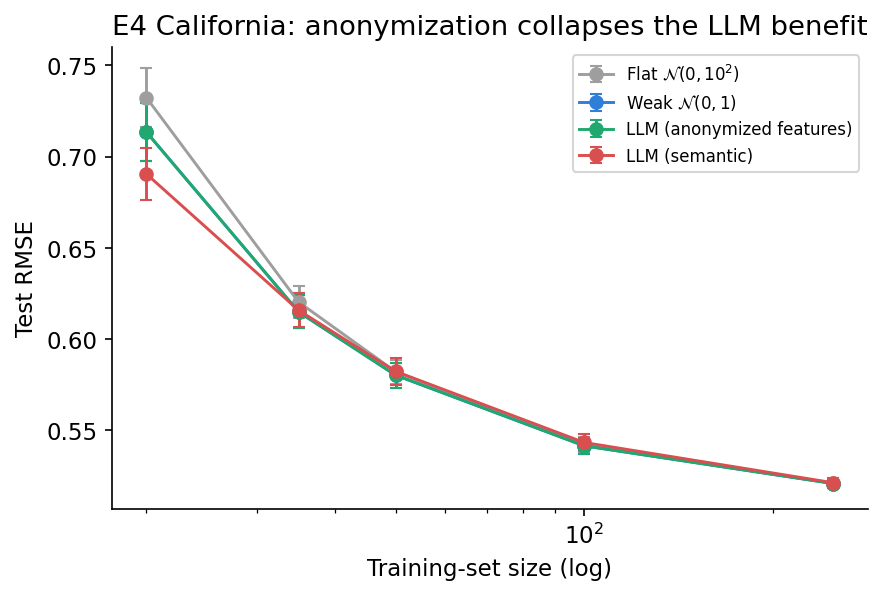}
\caption{E4 (\claude, semi-synthetic California): with feature names removed, the LLM prior does no better than the weak baseline at any $n$.}
\label{fig:e4}
\end{minipage}\hfill
\begin{minipage}[t]{0.48\textwidth}\centering
\includegraphics[width=0.8\linewidth]{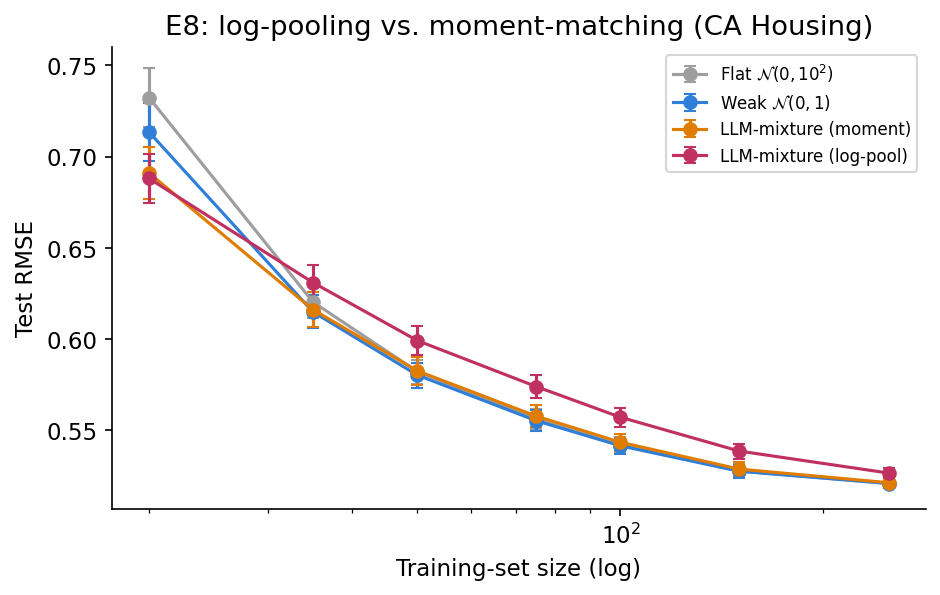}
\caption{E8: log pool vs.\ moment match of the $K{=}3$ \claude{} draws on semi-synthetic California.}
\label{fig:e8}
\end{minipage}
\end{figure*}

\section{Revision Analyses (T1): One-Protocol Re-Evaluation}\label{app:t1}

\begin{table}[tbp]\centering\small
\caption{Prior center $\times$ prior scale at $n{=}20$, RMSE (100 paired splits; \claude{} prior, $K{=}3$; NYC-2026: \gemini). Rows: center zero or $\mllm$. Columns: fixed scale ($\mathcal N(\bm0,\bm I)$ at zero, the elicited scales at $\mllm$), calibrated isotropic scale (EB-iso at zero), per-coordinate scale (ARD; EB-pc at zero). Two-vendor aggregates: Table~\ref{tab:rmse}. $^\dagger$Mean RMSE (the two-vendor log pool reaches $1.093$); other losses reverse the ranking (Table~\ref{tab:losses}).}
\label{tab:factorial}
\setlength{\tabcolsep}{4pt}
\resizebox{0.95\columnwidth}{!}{%
\begin{tabular}{llccc}
\toprule
Benchmark & Center & Fixed & Cal.\ iso & Per-coord \\
\midrule
\multirow{2}{*}{California} & $\bm 0$ & 0.714 & 0.730 & 0.715 \\
 & $\mllm$ & \textbf{0.691} & 0.708 & 0.712 \\
\midrule
\multirow{2}{*}{Diabetes} & $\bm 0$ & 0.972 & 0.852 & 0.893 \\
 & $\mllm$ & \textbf{0.846} & 0.850 & 0.897 \\
\midrule
\multirow{2}{*}{NYC-2026} & $\bm 0$ & 1.125 & 1.074 & 1.085 \\
 & $\mllm$ & \textbf{1.011} & 1.019 & 1.050 \\
\midrule
\multirow{2}{*}{California (real)$^\dagger$} & $\bm 0$ & 2.193 & 2.071 & \textbf{1.726} \\
 & $\mllm$ & 2.160 & 2.171 & 2.163 \\
\bottomrule
\end{tabular}}
\end{table}

\paragraph{Protocol and estimated $\sigma^2$.}
All tables in this section except Table~\ref{tab:t1stress} and the
stress and Breast Cancer rows of Table~\ref{tab:t1cross} come from a single seeded grid of 100 paired
random splits per $(\text{benchmark},n)$ (the R1 grid,
$\le1000$ held-out rows), regenerated from the recorded elicitations by
\texttt{tier1\_offline.py}; sources \texttt{T1\_*.csv}.
Matched comparisons on shared data are also how gains are separated when
automatic scores assist human evaluation \citep{liu2026human}.
``1v'' $=$ primary
vendor only ($K{=}3$); ``2v'' $=$ both vendors' draws pooled.  A
prior-agnostic $\hat\sigma^2$ changes no benchmark's top method
(Kendall $\tau$ $0.82$--$1.00$; Table~\ref{tab:t1sigma}).

\begin{table*}[tbp]\centering\footnotesize
\caption{Four losses at $n{=}35$, laid out as Table~\ref{tab:losses} (column groups: Calif.\ (semi), Diabetes, NYC-2026, Calif.\ (real); 1v $=$ \claude, 2v adds \gemini, NYC-2026: \gemini{} only).}\label{tab:t1losses35}
\resizebox{\textwidth}{!}{%
\begin{tabular}{lcccccccccccccccc}
\toprule
Method & RMSE & RMSE$_{\mathrm{trim}}$ & MedAE & $\|\bm m_n-\bm\beta^\star\|$ & RMSE & RMSE$_{\mathrm{trim}}$ & MedAE & $\|\bm m_n-\bm\beta^\star\|$ & RMSE & RMSE$_{\mathrm{trim}}$ & MedAE & $\|\bm m_n-\bm\beta^\star\|$ & RMSE & RMSE$_{\mathrm{trim}}$ & MedAE & $\|\bm m_n-\bm\beta^\star\|$ \\
\midrule
Weak $\mathcal N(\bm0,\bm I)$ & 0.615 & 0.597 & 0.417 & 0.542 & 0.850 & 0.831 & 0.590 & 0.851 & 0.961 & 0.889 & 0.517 & 0.539 & 2.030 & 0.671 & 0.416 & 0.607 \\
EB-iso & 0.624 & 0.604 & 0.424 & 0.639 & 0.802 & 0.798 & 0.581 & 0.724 & 0.948 & 0.874 & 0.486 & 0.464 & 1.974 & 0.690 & 0.427 & 0.753 \\
EB-pc (ARD) & 0.618 & 0.599 & 0.421 & 0.691 & 0.826 & 0.816 & 0.590 & 0.778 & 0.949 & 0.881 & 0.497 & 0.474 & 1.611 & 0.688 & 0.419 & 0.846 \\
LLM-mom (1v) & 0.616 & 0.598 & 0.418 & 0.636 & 0.799 & 0.796 & 0.564 & 0.683 & 0.916 & 0.877 & 0.478 & 0.372 & 2.008 & 0.668 & 0.411 & 0.684 \\
LLM-mom (2v) & 0.623 & 0.604 & 0.423 & 0.737 & 0.791 & 0.789 & 0.559 & 0.693 & 0.916 & 0.877 & 0.478 & 0.372 & 1.922 & 0.673 & 0.412 & 0.746 \\
LLM-cal & 0.638 & 0.617 & 0.432 & 0.821 & 0.795 & 0.792 & 0.567 & 0.683 & 0.913 & 0.870 & 0.486 & 0.360 & 2.022 & 0.692 & 0.433 & 0.868 \\
LLM-pc & 0.624 & 0.605 & 0.423 & 0.742 & 0.824 & 0.813 & 0.584 & 0.803 & 0.928 & 0.881 & 0.496 & 0.418 & 2.041 & 0.677 & 0.419 & 0.795 \\
LLM-pool (2v) & 0.667 & 0.643 & 0.448 & 1.101 & 0.798 & 0.793 & 0.574 & 0.678 & 0.942 & 0.905 & 0.507 & 0.407 & 1.036 & 0.730 & 0.451 & 0.982 \\
\bottomrule
\end{tabular}}
\end{table*}

\begin{table*}[tbp]\centering\footnotesize
\caption{Four losses at $n{=}50$, laid out as Table~\ref{tab:losses} (column groups: Calif.\ (semi), Diabetes, NYC-2026, Calif.\ (real); 1v $=$ \claude, 2v adds \gemini, NYC-2026: \gemini{} only).}\label{tab:t1losses50}
\resizebox{\textwidth}{!}{%
\begin{tabular}{lcccccccccccccccc}
\toprule
Method & RMSE & RMSE$_{\mathrm{trim}}$ & MedAE & $\|\bm m_n-\bm\beta^\star\|$ & RMSE & RMSE$_{\mathrm{trim}}$ & MedAE & $\|\bm m_n-\bm\beta^\star\|$ & RMSE & RMSE$_{\mathrm{trim}}$ & MedAE & $\|\bm m_n-\bm\beta^\star\|$ & RMSE & RMSE$_{\mathrm{trim}}$ & MedAE & $\|\bm m_n-\bm\beta^\star\|$ \\
\midrule
Weak $\mathcal N(\bm0,\bm I)$ & 0.580 & 0.573 & 0.397 & 0.469 & 0.799 & 0.791 & 0.555 & 0.762 & 0.906 & 0.849 & 0.488 & 0.463 & 1.887 & 0.636 & 0.381 & 0.521 \\
EB-iso & 0.585 & 0.576 & 0.400 & 0.527 & 0.772 & 0.770 & 0.555 & 0.706 & 0.900 & 0.842 & 0.466 & 0.418 & 1.865 & 0.645 & 0.387 & 0.631 \\
EB-pc (ARD) & 0.583 & 0.575 & 0.398 & 0.586 & 0.787 & 0.784 & 0.563 & 0.706 & 0.895 & 0.842 & 0.470 & 0.408 & 1.673 & 0.652 & 0.387 & 0.682 \\
LLM-mom (1v) & 0.583 & 0.574 & 0.398 & 0.548 & 0.771 & 0.769 & 0.546 & 0.667 & 0.876 & 0.845 & 0.454 & 0.341 & 1.898 & 0.637 & 0.380 & 0.612 \\
LLM-mom (2v) & 0.590 & 0.580 & 0.402 & 0.646 & 0.766 & 0.764 & 0.541 & 0.679 & 0.876 & 0.845 & 0.454 & 0.341 & 1.848 & 0.643 & 0.383 & 0.683 \\
LLM-cal & 0.599 & 0.588 & 0.406 & 0.678 & 0.766 & 0.765 & 0.546 & 0.669 & 0.870 & 0.832 & 0.460 & 0.319 & 1.925 & 0.662 & 0.402 & 0.832 \\
LLM-pc & 0.592 & 0.583 & 0.402 & 0.655 & 0.785 & 0.781 & 0.558 & 0.764 & 0.885 & 0.843 & 0.473 & 0.376 & 1.887 & 0.645 & 0.387 & 0.700 \\
LLM-pool (2v) & 0.647 & 0.631 & 0.434 & 1.089 & 0.763 & 0.761 & 0.549 & 0.672 & 0.904 & 0.876 & 0.485 & 0.389 & 1.022 & 0.710 & 0.432 & 0.974 \\
\bottomrule
\end{tabular}}
\end{table*}

\begin{table*}[tbp]\centering\footnotesize
\caption{Real California at all three sample sizes, with strong isotropic shrinkage controls: mean RMSE, trimmed RMSE, median AE, coefficient error and 90\% coefficient-interval coverage. At $n{=}20$, EB-pc minus $\mathcal N(\bm0,0.1\bm I)$ is $-0.31$ $[-0.63,-0.06]$ in mean RMSE but $-0.010$ $[-0.019,-0.002]$ in trimmed RMSE and $+0.09$ $[+0.03,+0.16]$ in coefficient error. Elicitors as in Table~\ref{tab:rmse}.}\label{tab:t1realca}
\resizebox{\textwidth}{!}{%
\begin{tabular}{l ccccc ccccc ccccc}
\toprule
 & \multicolumn{5}{c}{$n{=}20$} & \multicolumn{5}{c}{$n{=}35$} & \multicolumn{5}{c}{$n{=}50$} \\
\cmidrule(lr){2-6}\cmidrule(lr){7-11}\cmidrule(lr){12-16}
Prior & RMSE & RMSE$_{\mathrm{trim}}$ & MedAE & coef.\ err. & coef.\ cov. & RMSE & RMSE$_{\mathrm{trim}}$ & MedAE & coef.\ err. & coef.\ cov. & RMSE & RMSE$_{\mathrm{trim}}$ & MedAE & coef.\ err. & coef.\ cov. \\
\midrule
Flat & 2.262 & 0.714 & 0.510 & 1.099 & 0.89 & 2.061 & 0.675 & 0.422 & 0.746 & 0.85 & 1.901 & 0.639 & 0.385 & 0.608 & 0.83 \\
$\mathcal N(\bm0,4\bm I)$ & 2.223 & 0.708 & 0.499 & 0.883 & 0.89 & 2.050 & 0.673 & 0.419 & 0.686 & 0.85 & 1.896 & 0.638 & 0.383 & 0.574 & 0.83 \\
Weak $\mathcal N(\bm0,\bm I)$ & 2.193 & 0.705 & 0.487 & 0.753 & 0.90 & 2.030 & 0.671 & 0.416 & 0.607 & 0.86 & 1.887 & 0.636 & 0.381 & 0.521 & 0.84 \\
$\mathcal N(\bm0,0.25\bm I)$ & 2.150 & 0.713 & 0.488 & 0.845 & 0.79 & 1.999 & 0.677 & 0.417 & 0.648 & 0.80 & 1.873 & 0.639 & 0.382 & 0.550 & 0.78 \\
$\mathcal N(\bm0,0.1\bm I)$ & 2.039 & 0.732 & 0.501 & 0.983 & 0.59 & 1.953 & 0.697 & 0.432 & 0.810 & 0.58 & 1.857 & 0.653 & 0.393 & 0.709 & 0.58 \\
EB-iso & 2.071 & 0.731 & 0.504 & 0.978 & 0.59 & 1.974 & 0.690 & 0.427 & 0.753 & 0.65 & 1.865 & 0.645 & 0.387 & 0.631 & 0.68 \\
EB-pc (ARD) & 1.726 & 0.722 & 0.480 & 1.072 & 0.31 & 1.611 & 0.688 & 0.419 & 0.846 & 0.40 & 1.673 & 0.652 & 0.387 & 0.682 & 0.42 \\
CV ridge & 2.244 & 0.711 & 0.495 & 0.978 & 0.83 & 2.040 & 0.684 & 0.427 & 0.732 & 0.77 & 1.868 & 0.643 & 0.388 & 0.604 & 0.77 \\
LLM-fixed (1v) & 2.160 & 0.686 & 0.453 & 0.793 & 0.64 & 2.007 & 0.668 & 0.410 & 0.681 & 0.65 & 1.897 & 0.637 & 0.380 & 0.608 & 0.65 \\
LLM-mom (2v) & 2.005 & 0.689 & 0.449 & 0.835 & 0.57 & 1.922 & 0.673 & 0.412 & 0.746 & 0.54 & 1.848 & 0.643 & 0.383 & 0.683 & 0.53 \\
LLM-pool (2v) & 1.093 & 0.741 & 0.475 & 0.991 & 0.40 & 1.036 & 0.730 & 0.451 & 0.982 & 0.38 & 1.022 & 0.710 & 0.432 & 0.974 & 0.35 \\
LLM-cal & 2.171 & 0.707 & 0.469 & 0.923 & 0.33 & 2.022 & 0.692 & 0.433 & 0.868 & 0.39 & 1.925 & 0.662 & 0.402 & 0.832 & 0.41 \\
\bottomrule
\end{tabular}}
\end{table*}

\begin{table*}[tbp]\centering\footnotesize
\caption{Alternative prior centers, all with the same data-calibrated isotropic scale (the weak row keeps its fixed scale): displacement, RMSE and coefficient error at $n{=}20$, and the paired RMSE difference from the elicited center (95\% bootstrap CI, Wilcoxon $p$). Elicitors as in Table~\ref{tab:rmse}.}\label{tab:t1centers}
\resizebox{\textwidth}{!}{%
\begin{tabular}{llcccl}
\toprule
Benchmark & Center & $\delta$ & RMSE & coef.\ err. & $\Delta$ vs.\ $\mllm$ [CI] ($p$) \\
\midrule
Calif.\ (semi) & $\mllm$ (elicited) &  & 0.708 & 0.979 &  \\
Calif.\ (semi) & training-sample marginal slopes ($n{=}20$) & 1.31 & 0.770 & 1.048 & +0.062 [+0.050,+0.075] (7.9e-13) \\
Calif.\ (semi) & population marginal slopes (oracle) & 1.17 & 0.709 & 1.089 & +0.001 [$-$0.004,+0.007] (3.2e-01) \\
Calif.\ (semi) & $(\widehat{\bm R}_{20}+0.1\bm I)^{-1}\mllm$ & 0.89 & 0.718 & 0.736 & +0.010 [$-$0.008,+0.030] (7.3e-01) \\
Calif.\ (semi) & $\bm R_{\mathrm{full}}^{-1}\mllm$ & 2.34 & 0.818 & 1.618 & +0.111 [+0.087,+0.138] (6.0e-12) \\
Calif.\ (semi) & elicited signs $\times0.3$ & 1.30 & 0.735 & 0.953 & +0.028 [+0.017,+0.038] (1.9e-06) \\
Calif.\ (semi) & zero (EB-iso) &  & 0.730 & 0.888 & +0.023 [+0.010,+0.036] (2.3e-03) \\
Calif.\ (semi) & zero (weak) &  & 0.714 & 0.684 & +0.006 [$-$0.010,+0.023] (7.8e-01) \\
Diabetes & $\mllm$ (elicited) &  & 0.850 & 0.703 &  \\
Diabetes & training-sample marginal slopes ($n{=}20$) & 1.22 & 0.912 & 0.814 & +0.062 [+0.052,+0.073] (5.3e-15) \\
Diabetes & population marginal slopes (oracle) & 1.07 & 0.886 & 0.778 & +0.036 [+0.028,+0.045] (7.6e-12) \\
Diabetes & $(\widehat{\bm R}_{20}+0.1\bm I)^{-1}\mllm$ & 0.73 & 0.875 & 0.681 & +0.025 [+0.011,+0.039] (4.2e-04) \\
Diabetes & $\bm R_{\mathrm{full}}^{-1}\mllm$ & 3.16 & 0.957 & 3.101 & +0.107 [+0.092,+0.123] (8.0e-17) \\
Diabetes & elicited signs $\times0.3$ & 1.00 & 0.884 & 0.805 & +0.034 [+0.028,+0.040] (1.2e-15) \\
Diabetes & zero (EB-iso) &  & 0.852 & 0.747 & +0.002 [$-$0.008,+0.012] (2.7e-01) \\
Diabetes & zero (weak) &  & 0.972 & 0.983 & +0.122 [+0.102,+0.141] (1.0e-16) \\
NYC-2026 & $\mllm$ (elicited) &  & 1.019 & 0.414 &  \\
NYC-2026 & training-sample marginal slopes ($n{=}20$) & 0.78 & 1.093 & 0.581 & +0.073 [+0.057,+0.090] (1.1e-11) \\
NYC-2026 & population marginal slopes (oracle) & 0.49 & 1.009 & 0.430 & $-$0.011 [$-$0.019,$-$0.002] (2.2e-02) \\
NYC-2026 & $(\widehat{\bm R}_{20}+0.1\bm I)^{-1}\mllm$ & 4.04 & 3.534 & 3.983 & +2.514 [+2.285,+2.716] (3.9e-18) \\
NYC-2026 & $\bm R_{\mathrm{full}}^{-1}\mllm$ & 3.38 & 1.352 & 2.182 & +0.333 [+0.302,+0.369] (4.7e-18) \\
NYC-2026 & elicited signs $\times0.3$ & 0.73 & 1.057 & 0.553 & +0.038 [+0.026,+0.051] (3.0e-07) \\
NYC-2026 & zero (EB-iso) &  & 1.074 & 0.531 & +0.055 [+0.035,+0.074] (1.6e-06) \\
NYC-2026 & zero (weak) &  & 1.125 & 0.691 & +0.106 [+0.086,+0.128] (3.3e-14) \\
Calif.\ (real) & $\mllm$ (elicited) &  & 2.171 & 0.923 &  \\
Calif.\ (real) & training-sample marginal slopes ($n{=}20$) & 1.30 & 2.457 & 1.044 & +0.285 [+0.001,+0.580] (1.2e-08) \\
Calif.\ (real) & population marginal slopes (oracle) & 1.07 & 1.460 & 1.014 & $-$0.712 [$-$1.186,$-$0.321] (8.5e-02) \\
Calif.\ (real) & $(\widehat{\bm R}_{20}+0.1\bm I)^{-1}\mllm$ & 0.98 & 2.441 & 0.883 & +0.270 [$-$0.366,+0.939] (9.0e-07) \\
Calif.\ (real) & $\bm R_{\mathrm{full}}^{-1}\mllm$ & 2.92 & 2.602 & 2.348 & +0.431 [+0.066,+0.843] (7.6e-09) \\
Calif.\ (real) & elicited signs $\times0.3$ & 1.19 & 2.514 & 0.922 & +0.343 [+0.151,+0.576] (2.5e-10) \\
Calif.\ (real) & zero (EB-iso) &  & 2.071 & 0.978 & $-$0.100 [$-$0.320,+0.101] (1.1e-03) \\
Calif.\ (real) & zero (weak) &  & 2.193 & 0.753 & +0.022 [$-$0.207,+0.252] (2.2e-04) \\
\bottomrule
\end{tabular}}
\end{table*}

\begin{table}[htbp]\centering\footnotesize
\caption{Does estimating $\sigma^2$ change the ranking? Kendall $\tau$ between the $n{=}20$ method rankings under the oracle and a prior-agnostic estimated $\sigma^2$, the top method under each, the best zero-centered prior under the estimate, and the LLM-fixed margin against it. Elicitors as in Table~\ref{tab:rmse}.}\label{tab:t1sigma}
\resizebox{\columnwidth}{!}{%
\begin{tabular}{lccccc}
\toprule
Benchmark & Kendall $\tau$ & Top (oracle) & Top (est.) & Best classical (est.) & LLM-fixed vs.\ best cls. \\
\midrule
Calif.\ (semi) & 0.96 & LLM-mom (2v) & LLM-mom (2v) & Weak $\mathcal N(\bm0,\bm I)$ & $-$3.2\% \\
Diabetes & 1.00 & LLM-mom (2v) & LLM-mom (2v) & EB-iso & $-$4.8\% \\
NYC-2026 & 0.93 & LLM-mom (1v) & LLM-mom (1v) & EB-pc (ARD) & $-$7.6\% \\
Calif.\ (real) & 0.82 & LLM-pool (2v) & LLM-pool (2v) & EB-pc (ARD) & +23.9\% \\
\bottomrule
\end{tabular}}
\end{table}

\begin{table*}[tbp]\centering\footnotesize
\caption{Share of the $p$ coefficients covered by their 90\% credible intervals at $n{=}20$ and $50$ (averaged over splits). Elicitors as in Table~\ref{tab:rmse}.}\label{tab:t1coefcov}
\resizebox{\textwidth}{!}{%
\begin{tabular}{lcccccccc}
\toprule
Method & semi-CA 20 & semi-CA 50 & Diab.\ 20 & Diab.\ 50 & NYC 20 & NYC 50 & real-CA 20 & real-CA 50 \\
\midrule
Flat & 0.89 & 0.88 & 0.90 & 0.91 & 0.90 & 0.89 & 0.89 & 0.83 \\
Weak $\mathcal N(\bm0,\bm I)$ & 0.90 & 0.88 & 0.94 & 0.93 & 0.93 & 0.90 & 0.90 & 0.84 \\
EB-iso & 0.69 & 0.80 & 0.73 & 0.66 & 0.83 & 0.83 & 0.59 & 0.68 \\
EB-pc (ARD) & 0.34 & 0.41 & 0.25 & 0.29 & 0.38 & 0.45 & 0.31 & 0.42 \\
CV ridge & 0.83 & 0.87 & 0.80 & 0.73 & 0.77 & 0.82 & 0.83 & 0.77 \\
LLM-fixed (1v) & 0.66 & 0.73 & 0.81 & 0.75 & 0.60 & 0.62 & 0.64 & 0.65 \\
LLM-mom (1v) & 0.66 & 0.73 & 0.81 & 0.75 & 0.62 & 0.65 & 0.63 & 0.64 \\
LLM-mom (2v) & 0.55 & 0.60 & 0.79 & 0.72 & 0.62 & 0.65 & 0.57 & 0.53 \\
LLM-pool (2v) & 0.38 & 0.38 & 0.50 & 0.52 & 0.39 & 0.45 & 0.40 & 0.35 \\
LLM-cal & 0.42 & 0.60 & 0.69 & 0.63 & 0.71 & 0.85 & 0.33 & 0.41 \\
LLM-pc & 0.25 & 0.38 & 0.19 & 0.25 & 0.34 & 0.46 & 0.16 & 0.27 \\
\bottomrule
\end{tabular}}
\end{table*}

\begin{table*}[tbp]\centering\footnotesize
\caption{Stress family (ten DGP seeds $\times$ 50 paired splits; \claude{} prior, $K{=}3$): held-out log-loss of the fixed, calibrated and robust-mixture priors above the comparator $\mathcal N(\bm0,\tau_0^2\bm I)$ (95\% bootstrap CI; in parentheses, how many of the ten seeds show harm). The truth is drawn from $\mathcal N(\bm0,0.36\bm I)$, so $\tau_0^2\in\{0.25,1\}$ is a near-oracle comparator; the mixture's vague component uses the same $\tau_0^2$.}\label{tab:t1stress}
\resizebox{\textwidth}{!}{%
\begin{tabular}{cccccc}
\toprule
$n$ & $\tau_0^2$ & weak & fixed & calibrated & mixture \\
\midrule
20 & 0.1 & 0.637 & +0.088 [+0.0812,+0.0947] (10) & +0.016 [+0.0110,+0.0222] (8) & $-$0.002 [$-$0.0043,+0.0012] (6) \\
20 & 0.25 & 0.624 & +0.101 [+0.0938,+0.1085] (10) & +0.029 [+0.0248,+0.0344] (10) & +0.008 [+0.0056,+0.0107] (9) \\
20 & 1 & 0.630 & +0.095 [+0.0865,+0.1039] (9) & +0.023 [+0.0188,+0.0283] (9) & +0.010 [+0.0068,+0.0134] (9) \\
20 & 4 & 0.657 & +0.068 [+0.0579,+0.0778] (7) & $-$0.004 [$-$0.0109,+0.0020] (4) & $-$0.003 [$-$0.0087,+0.0024] (4) \\
35 & 0.1 & 0.614 & +0.059 [+0.0546,+0.0638] (9) & $-$0.002 [$-$0.0052,+0.0022] (6) & $-$0.005 [$-$0.0078,$-$0.0029] (6) \\
35 & 0.25 & 0.599 & +0.074 [+0.0686,+0.0790] (10) & +0.013 [+0.0100,+0.0159] (8) & +0.005 [+0.0028,+0.0065] (8) \\
35 & 1 & 0.602 & +0.071 [+0.0646,+0.0778] (9) & +0.010 [+0.0073,+0.0132] (9) & +0.005 [+0.0031,+0.0078] (9) \\
35 & 4 & 0.617 & +0.056 [+0.0487,+0.0632] (8) & $-$0.005 [$-$0.0088,$-$0.0011] (4) & $-$0.004 [$-$0.0071,+0.0001] (4) \\
50 & 0.1 & 0.599 & +0.045 [+0.0417,+0.0487] (9) & $-$0.007 [$-$0.0096,$-$0.0037] (5) & $-$0.007 [$-$0.0089,$-$0.0045] (4) \\
50 & 0.25 & 0.585 & +0.059 [+0.0548,+0.0632] (10) & +0.007 [+0.0050,+0.0092] (8) & +0.003 [+0.0013,+0.0044] (8) \\
50 & 1 & 0.587 & +0.057 [+0.0519,+0.0623] (9) & +0.005 [+0.0035,+0.0072] (9) & +0.003 [+0.0015,+0.0044] (9) \\
50 & 4 & 0.597 & +0.047 [+0.0414,+0.0528] (8) & $-$0.005 [$-$0.0072,$-$0.0025] (2) & $-$0.004 [$-$0.0058,$-$0.0016] (2) \\
\bottomrule
\end{tabular}}
\end{table*}

\begin{table*}[tbp]\centering\footnotesize
\caption{Cross-problem summary at $n{=}20$: paired change relative to the best zero-centered prior (95\% bootstrap CI) and the standardized effect $d=\text{mean}/\text{SD}$ over splits. The single-vendor fixed prior improves on the best classical method on 4 of 6 problems (sign test $p=0.69$). Elicitors as in Table~\ref{tab:rmse}; these CIs are an independent bootstrap replicate and can differ from the text in the last digit.}\label{tab:t1cross}
\resizebox{\textwidth}{!}{%
\begin{tabular}{llllcc}
\toprule
Problem & Loss & Method & vs. & $\Delta$ [CI] & $d$ \\
\midrule
Calif.\ (semi) & RMSE & LLM-fixed (1v) & Weak $\mathcal N(\bm0,\bm I)$ & $-$3.2\% [$-$5.0,$-$1.5] & $-$0.34 \\
Calif.\ (semi) & RMSE & LLM-mom (2v) & Weak $\mathcal N(\bm0,\bm I)$ & $-$3.2\% [$-$5.5,$-$1.1] & $-$0.29 \\
Calif.\ (semi) & RMSE & LLM-cal & Weak $\mathcal N(\bm0,\bm I)$ & $-$0.8\% [$-$3.3,+1.5] & $-$0.07 \\
Diabetes & RMSE & LLM-fixed (1v) & EB-iso & $-$0.8\% [$-$1.8,+0.3] & $-$0.14 \\
Diabetes & RMSE & LLM-mom (2v) & EB-iso & $-$2.2\% [$-$3.5,$-$1.0] & $-$0.35 \\
Diabetes & RMSE & LLM-cal & EB-iso & $-$0.3\% [$-$1.4,+0.9] & $-$0.04 \\
NYC-2026 & RMSE & LLM-fixed (1v) & EB-iso & $-$5.8\% [$-$8.0,$-$3.5] & $-$0.50 \\
NYC-2026 & RMSE & LLM-mom (2v) & EB-iso & $-$5.8\% [$-$8.0,$-$3.6] & $-$0.50 \\
NYC-2026 & RMSE & LLM-cal & EB-iso & $-$5.1\% [$-$7.0,$-$3.1] & $-$0.54 \\
Calif.\ (real) & RMSE & LLM-fixed (1v) & EB-pc (ARD) & +25.1\% [+5.5,+48.2] & +0.23 \\
Calif.\ (real) & RMSE & LLM-mom (2v) & EB-pc (ARD) & +16.2\% [$-$3.1,+37.9] & +0.15 \\
Calif.\ (real) & RMSE & LLM-cal & EB-pc (ARD) & +25.8\% [+2.8,+53.6] & +0.20 \\
stress (logistic, 10 seeds) & log-loss & LLM-fixed (1v) & Weak $\mathcal N(\bm0,\bm I)$ & +15.1\% [+13.8,+16.5] & +0.96 \\
stress (logistic, 10 seeds) & log-loss & LLM-cal & Weak $\mathcal N(\bm0,\bm I)$ & +3.7\% [+3.0,+4.5] & +0.42 \\
Breast Cancer (E16) & log-loss & Claude semantic & Weak $\mathcal N(\bm0,\bm I)$ & $-11.5\%$ & --- \\
Breast Cancer (E16) & log-loss & Gemini semantic & Weak $\mathcal N(\bm0,\bm I)$ & $+15.9\%$ & --- \\
\bottomrule
\end{tabular}}
\end{table*}

\providecommand{\bm}{\boldsymbol}
\providecommand{\rtfitcol}[1]{\resizebox{\ifdim\width>\columnwidth\columnwidth\else\width\fi}{!}{#1}}
\providecommand{\rtfitwide}[1]{\resizebox{\ifdim\width>\textwidth\textwidth\else\width\fi}{!}{#1}}
\makeatletter
\providecommand{\fps@rtab}{htbp}\providecommand{\ftype@rtab}{4}
\newenvironment{rtab}[1]%
  {\@float{rtab}[htbp]\centering\footnotesize
   {\raggedright\scriptsize\textbf{#1}\par\vskip 1pt}}%
  {\end@float}
\makeatother

\section{Supplementary Tables}\label{app:rtables}
\edef\restoreintextsep{\intextsep=\the\intextsep\relax}%
\setlength{\intextsep}{4pt plus 2pt minus 1pt}%
\renewcommand{\textfraction}{0.15}%

Tables in this appendix keep the R-numbers of the released result
files they render, R3 to R27 with gaps, and the text cites them by
these numbers; all other tables use the running counter.  The files,
CSV or JSON, are in
\texttt{results/major\_\allowbreak{}revision/processed/}; a letter
separates files with the same number.  A table can be placed away
from its heading; the tag above it names it.
Conventions: floats are rounded to three
decimals; $p$-values below $0.001$ are shown with two significant digits in
scientific notation and an exact $0$ denotes a $p$-value below the
numerical resolution of the test; booleans are Y/N; ``--'' marks a missing
entry; method and benchmark identifiers and verdict strings are printed exactly as in the source
files (glossary: Appendix~\ref{app:gloss}; ``registered'' in a verdict means specified before the elicitation, not lodged with a registry), and entries can differ from the T1 tables in the last digit.
\texttt{california} is the semi-synthetic benchmark,
\texttt{calif\_real} the real one and \texttt{logistic} the
stress test.  Long per-split files are aggregated over splits (means) and
the aggregation is stated in the description.

\subsection*{Table~R3: Error-level dependence of elicitations (E11)}{\makeatletter\def\@currentlabel{R3}\makeatother\label{tab:R3}}
Per benchmark: number of elicitor models and recorded draws, mean pairwise correlation of the raw elicited mean vectors and of the error vectors $\bm e_k=\bm m_k-\bm\beta^\star$ (immune to shared-signal inflation), the within-vendor intraclass correlation of errors beyond the common component, and the variance decomposition of the elicitations into shared bias, draw noise and signal (source \texttt{R3\_\allowbreak{}error\_\allowbreak{}dependence.csv}). (Full-width table, floated.)
\begin{table*}[t]\centering\footnotesize\setlength{\tabcolsep}{3.5pt}
{\raggedright\textbf{Table~\ref{tab:R3}:} Error-level dependence of elicitations (E11).\par}\smallskip
\rtfitwide{%
\begin{tabular}{lrrrrrrrr}
\toprule
benchmark & models & draws & corr(raw) & corr(err.) & ICC(err.) & shared-bias var. & draw-noise var. & signal var. \\
\midrule
california & 2 & 6 & 0.977 & 0.984 & 0.160 & 0.001 & 0.002 & 0.268 \\
diabetes & 2 & 6 & 0.922 & 0.951 & 0.437 & 0.003 & 0.001 & 0.065 \\
nyc2026 & 1 & 3 & 0.980 & 0.960 & -- & 0.000 & 0.001 & 0.033 \\
logistic & 2 & 6 & 0.988 & 0.931 & 0.913 & 0.025 & 0.002 & 0.222 \\
bcancer & 2 & 6 & 0.889 & 0.989 & 0.790 & 0.021 & 0.003 & 0.598 \\
\bottomrule
\end{tabular}}
\end{table*}

\subsection*{Table~\ref{tab:R4}a: Mean held-out RMSE per prior and cell, R1 factorial (Gaussian benchmarks)}{\makeatletter\def\@currentlabel{R4}\makeatother\label{tab:R4}}
Mean predictive RMSE over $100$ paired splits for every prior, benchmark and training size of the R1 factorial; bold marks the best prior in each cell (\texttt{is\_best}), so a prior's regret in a cell is its entry minus the bold entry (source \texttt{R4\_\allowbreak{}regret\_\allowbreak{}R1.csv}). (Full-width table, floated.)
\begin{table*}[t]\centering\footnotesize\setlength{\tabcolsep}{3.5pt}
{\raggedright\textbf{Table~\ref{tab:R4}a:} Mean held-out RMSE per prior and cell, R1 factorial (Gaussian benchmarks).\par}\smallskip
\rtfitwide{%
\begin{tabular}{lrrrrrrrrrrrrrrrr}
\toprule
 & \multicolumn{4}{c}{calif\_real} & \multicolumn{4}{c}{california} & \multicolumn{4}{c}{diabetes} & \multicolumn{4}{c}{nyc2026} \\
\cmidrule(lr){2-5} \cmidrule(lr){6-9} \cmidrule(lr){10-13} \cmidrule(lr){14-17}
prior & $n{=}20$ & $n{=}35$ & $n{=}50$ & $n{=}100$ & $n{=}20$ & $n{=}35$ & $n{=}50$ & $n{=}100$ & $n{=}20$ & $n{=}35$ & $n{=}50$ & $n{=}100$ & $n{=}20$ & $n{=}35$ & $n{=}50$ & $n{=}100$ \\
\midrule
cv\_ridge & 2.244 & 2.040 & 1.868 & 1.739 & 0.736 & 0.621 & 0.583 & 0.542 & 0.926 & 0.833 & 0.783 & 0.741 & 1.123 & 0.961 & 0.909 & 0.833 \\
eb\_zero\_iso & 2.071 & 1.974 & 1.866 & 1.735 & 0.730 & 0.624 & 0.585 & 0.543 & 0.852 & 0.802 & 0.772 & 0.738 & 1.074 & 0.948 & 0.900 & 0.830 \\
eb\_zero\_pc & 1.726 & 1.611 & 1.673 & 1.637 & 0.715 & 0.618 & 0.583 & 0.543 & 0.893 & 0.826 & 0.787 & 0.745 & 1.085 & 0.949 & 0.895 & 0.830 \\
flat & 2.262 & 2.061 & 1.901 & 1.745 & 0.732 & 0.620 & 0.582 & 0.542 & 1.151 & 0.906 & 0.820 & 0.749 & 1.159 & 0.970 & 0.910 & 0.832 \\
llm\_cal\_iso & 2.172 & 2.022 & 1.925 & 1.771 & 0.708 & 0.638 & 0.599 & 0.547 & 0.850 & 0.795 & 0.766 & 0.736 & 1.019 & \textbf{0.913} & 0.870 & 0.820 \\
llm\_cal\_pc & 2.163 & 2.041 & 1.887 & 1.750 & 0.712 & 0.624 & 0.592 & 0.546 & 0.897 & 0.824 & 0.785 & 0.745 & 1.050 & 0.928 & 0.885 & 0.829 \\
llm\_fixed & 2.160 & 2.007 & 1.897 & 1.760 & 0.691 & 0.616 & 0.582 & 0.543 & 0.846 & 0.800 & 0.771 & 0.738 & 1.011 & 0.916 & 0.876 & 0.827 \\
llm\_hier & 1.471 & 1.436 & 1.458 & 1.516 & 0.707 & 0.669 & 0.650 & 0.620 & 0.854 & \textbf{0.791} & \textbf{0.761} & 0.731 & \textbf{1.011} & 0.916 & 0.876 & 0.828 \\
llm\_logpool & \textbf{1.093} & \textbf{1.036} & \textbf{1.022} & \textbf{0.999} & 0.704 & 0.667 & 0.647 & 0.618 & 0.866 & 0.798 & 0.763 & \textbf{0.731} & 1.034 & 0.942 & 0.904 & 0.858 \\
llm\_moment & 2.005 & 1.922 & 1.848 & 1.745 & \textbf{0.690} & 0.623 & 0.590 & 0.548 & \textbf{0.834} & 0.792 & 0.766 & 0.736 & \textbf{1.011} & 0.916 & 0.876 & 0.828 \\
llm\_robust\_mix & 2.171 & 2.023 & 1.925 & 1.771 & 0.707 & 0.638 & 0.599 & 0.547 & 0.850 & 0.795 & 0.766 & 0.736 & 1.020 & 0.913 & \textbf{0.870} & \textbf{0.820} \\
weak\_zero\_iso & 2.193 & 2.030 & 1.887 & 1.741 & 0.713 & \textbf{0.615} & \textbf{0.580} & \textbf{0.542} & 0.972 & 0.850 & 0.799 & 0.745 & 1.125 & 0.961 & 0.906 & 0.831 \\
\bottomrule
\end{tabular}}
\end{table*}

\subsection*{Table~\ref{tab:R4}g: Paired tests against the weak prior, R2 misaligned stress family}
Mean paired difference in held-out log-loss relative to the weak prior on the seeded logistic stress family (positive = harm), its standard error over the $230$ paired (seed, split) evaluations per cell, and raw and BH-adjusted $p$-values, by training size (source \texttt{R4\_\allowbreak{}tests\_\allowbreak{}R2.csv}).
\begin{rtab}{Table~\ref{tab:R4}g}\setlength{\tabcolsep}{3.5pt}
\rtfitcol{%
\begin{tabular}{rlrrrr}
\toprule
$n$ & prior & $\Delta$ vs.\ weak & SE & $p_{\mathrm{raw}}$ & $p_{\mathrm{BH}}$ \\
\midrule
20 & flat & 0.068 & 0.005 & $5.1\times10^{-36}$ & $3.6\times10^{-35}$ \\
20 & llm\_cal\_iso & 0.023 & 0.004 & $3.6\times10^{-8}$ & $4.2\times10^{-8}$ \\
20 & llm\_fixed & 0.081 & 0.007 & $6.7\times10^{-24}$ & $1.2\times10^{-23}$ \\
20 & llm\_hier & 0.100 & 0.008 & $3.7\times10^{-25}$ & $8.7\times10^{-25}$ \\
20 & llm\_logpool & 0.111 & 0.008 & $2.7\times10^{-26}$ & $9.4\times10^{-26}$ \\
20 & llm\_moment & 0.046 & 0.005 & $2.3\times10^{-15}$ & $3.2\times10^{-15}$ \\
20 & llm\_robust\_mix & 0.011 & 0.003 & 0.028 & 0.028 \\
\midrule
35 & flat & 0.028 & 0.002 & $7.8\times10^{-31}$ & $2.7\times10^{-30}$ \\
35 & llm\_cal\_iso & 0.009 & 0.002 & $2.7\times10^{-4}$ & $3.1\times10^{-4}$ \\
35 & llm\_fixed & 0.060 & 0.005 & $2.4\times10^{-23}$ & $4.2\times10^{-23}$ \\
35 & llm\_hier & 0.102 & 0.007 & $1.3\times10^{-29}$ & $3.0\times10^{-29}$ \\
35 & llm\_logpool & 0.126 & 0.008 & $2.7\times10^{-32}$ & $1.9\times10^{-31}$ \\
35 & llm\_moment & 0.040 & 0.004 & $4.6\times10^{-18}$ & $6.4\times10^{-18}$ \\
35 & llm\_robust\_mix & 0.005 & 0.002 & 0.147 & 0.147 \\
\midrule
50 & flat & 0.018 & 0.002 & $4.2\times10^{-29}$ & $9.7\times10^{-29}$ \\
50 & llm\_cal\_iso & 0.006 & 0.002 & 0.002 & 0.003 \\
50 & llm\_fixed & 0.049 & 0.004 & $2.5\times10^{-23}$ & $4.3\times10^{-23}$ \\
50 & llm\_hier & 0.100 & 0.006 & $1.4\times10^{-32}$ & $4.8\times10^{-32}$ \\
50 & llm\_logpool & 0.132 & 0.008 & $2.2\times10^{-35}$ & $1.6\times10^{-34}$ \\
50 & llm\_moment & 0.036 & 0.003 & $3.1\times10^{-20}$ & $4.3\times10^{-20}$ \\
50 & llm\_robust\_mix & 0.003 & 0.001 & 0.215 & 0.215 \\
\midrule
100 & flat & 0.005 & 0.001 & $8.3\times10^{-18}$ & $1.2\times10^{-17}$ \\
100 & llm\_cal\_iso & 0.001 & 0.001 & 0.305 & 0.356 \\
100 & llm\_fixed & 0.030 & 0.002 & $1.5\times10^{-25}$ & $3.4\times10^{-25}$ \\
100 & llm\_hier & 0.091 & 0.005 & $5.6\times10^{-37}$ & $1.9\times10^{-36}$ \\
100 & llm\_logpool & 0.138 & 0.007 & $8.2\times10^{-39}$ & $5.8\times10^{-38}$ \\
100 & llm\_moment & 0.028 & 0.002 & $1.9\times10^{-25}$ & $3.4\times10^{-25}$ \\
100 & llm\_robust\_mix & 0.000 & 0.000 & 0.777 & 0.777 \\
\bottomrule
\end{tabular}}
\end{rtab}

\subsection*{Table~R7: Permuted-description probe: gate behavior and pilot verdict}{\makeatletter\def\@currentlabel{R7}\makeatother\label{tab:R7}}
Key/value dump of the two probe JSON files: the elicited mean vector of the adversarial permuted-description prior (\texttt{m\_perm}, $p{=}8$ coordinates), the trust rate of the matched-scale gate for that prior by training size, and the noise variance $\sigma^2$ (\texttt{R7\_\allowbreak{}probe\_\allowbreak{}gate.json}); and the pre-specified permutation-test statistics of the underpowered pilot, namely the correlation of the elicited mean with the permuted descriptions ($r_{\mathrm{perm}}$) vs.\ with the true columns ($r_{\mathrm{id}}$), their difference $D$, two-sided $p$ and verdict (\texttt{R7\_\allowbreak{}probe\_\allowbreak{}verdict.json}); the powered second stage is Table~R21.
\begin{rtab}{Table~R7}\setlength{\tabcolsep}{3.5pt}
\rtfitcol{%
\begin{tabular}{lp{1.7in}}
\toprule
key & value \\
\midrule
\multicolumn{2}{l}{\textit{\texttt{R7\_\allowbreak{}probe\_\allowbreak{}gate.json}}} \\
m\_perm[0] & 0.333 \\
m\_perm[1] & 0.167 \\
m\_perm[2] & $-0.083$ \\
m\_perm[3] & 0.600 \\
m\_perm[4] & 0.150 \\
m\_perm[5] & $-0.167$ \\
m\_perm[6] & 0.217 \\
m\_perm[7] & 0.600 \\
trust\_by\_n.20 & 0.025 \\
trust\_by\_n.35 & 0.000 \\
trust\_by\_n.50 & 0.000 \\
trust\_by\_n.100 & 0.000 \\
sigma2 & 0.552 \\
\midrule
\multicolumn{2}{l}{\textit{\texttt{R7\_\allowbreak{}probe\_\allowbreak{}verdict.json}}} \\
r\_perm & 0.960 \\
r\_id & 0.138 \\
D & 0.822 \\
p\_two & 0.129 \\
degenerate & N \\
verdict & INCONCLUSIVE under the pre-registered permutation test \\
\bottomrule
\end{tabular}}
\end{rtab}

\subsection*{Table~\ref{tab:R9}a: Gate confusion by benchmark: trust rates}{\makeatletter\def\@currentlabel{R9}\makeatother\label{tab:R9}}
Matched-scale trust gate ($\|\mllm-\widehat{\bm\beta}_{\mathrm{weak}}\|\le\|\widehat{\bm\beta}_{\mathrm{weak}}\|$) evaluated over $100$ paired splits per benchmark and training size ($50$ for \texttt{logistic\_stress}, the single DGP under which the elicitation was recorded): whether the cell is aligned (moment-matching beats weak in mean loss), the fraction of splits in which the gate trusts the elicited prior, and the fraction of splits in which its decision is correct (source \texttt{R9\_\allowbreak{}gate\_\allowbreak{}confusion.csv}).
\begin{rtab}{Table~\ref{tab:R9}a}\setlength{\tabcolsep}{3.5pt}
\rtfitcol{%
\begin{tabular}{lrrrr}
\toprule
benchmark & $n$ & aligned & trust rate & correct rate \\
\midrule
calif\_real & 20 & Y & 1.000 & 1.000 \\
calif\_real & 35 & Y & 1.000 & 1.000 \\
calif\_real & 50 & Y & 1.000 & 1.000 \\
calif\_real & 100 & N & 1.000 & 0.000 \\
\midrule
california & 20 & Y & 0.990 & 0.990 \\
california & 35 & N & 1.000 & 0.000 \\
california & 50 & N & 1.000 & 0.000 \\
california & 100 & N & 1.000 & 0.000 \\
\midrule
diabetes & 20 & Y & 0.830 & 0.830 \\
diabetes & 35 & Y & 0.860 & 0.860 \\
diabetes & 50 & Y & 0.900 & 0.900 \\
diabetes & 100 & Y & 0.990 & 0.990 \\
\midrule
logistic\_stress & 20 & Y & 0.520 & 0.520 \\
logistic\_stress & 35 & Y & 0.620 & 0.620 \\
logistic\_stress & 50 & Y & 0.620 & 0.620 \\
logistic\_stress & 100 & Y & 0.720 & 0.720 \\
\midrule
nyc2026 & 20 & Y & 0.450 & 0.450 \\
nyc2026 & 35 & Y & 0.620 & 0.620 \\
nyc2026 & 50 & Y & 0.680 & 0.680 \\
nyc2026 & 100 & Y & 0.830 & 0.830 \\
\bottomrule
\end{tabular}}
\end{rtab}

\subsection*{Table~\ref{tab:R9}b: Gate confusion on the misaligned stress family, by seed}
Same gate evaluated on the seeded logistic stress family (the nine DGP seeds drawn independently of the elicitation, $15$ paired splits each, $n\in\{20,50\}$): alignment of the cell, trust rate and correct-decision rate; $17$ of the $18$ cells are misaligned and the gate distrusts them in $91\%$ of splits (source \texttt{R9\_\allowbreak{}logistic\_\allowbreak{}confusion.csv}).
\begin{rtab}{Table~\ref{tab:R9}b}\setlength{\tabcolsep}{3.5pt}
\rtfitcol{%
\begin{tabular}{rrrrr}
\toprule
seed & $n$ & aligned & trust rate & correct rate \\
\midrule
12 & 20 & N & 0.200 & 0.800 \\
12 & 50 & N & 0.067 & 0.933 \\
\midrule
13 & 20 & N & 0.000 & 1.000 \\
13 & 50 & N & 0.000 & 1.000 \\
\midrule
14 & 20 & N & 0.000 & 1.000 \\
14 & 50 & N & 0.000 & 1.000 \\
\midrule
15 & 20 & N & 0.000 & 1.000 \\
15 & 50 & N & 0.000 & 1.000 \\
\midrule
16 & 20 & N & 0.200 & 0.800 \\
16 & 50 & N & 0.000 & 1.000 \\
\midrule
17 & 20 & N & 0.267 & 0.733 \\
17 & 50 & N & 0.200 & 0.800 \\
\midrule
18 & 20 & N & 0.000 & 1.000 \\
18 & 50 & N & 0.000 & 1.000 \\
\midrule
19 & 20 & N & 0.133 & 0.867 \\
19 & 50 & N & 0.067 & 0.933 \\
\midrule
20 & 20 & Y & 0.400 & 0.400 \\
20 & 50 & N & 0.400 & 0.600 \\
\bottomrule
\end{tabular}}
\end{rtab}

\subsection*{Table~R12: Paraphrase-level vs.\ same-prompt error correlations (E17)}{\makeatletter\def\@currentlabel{R12}\makeatother\label{tab:R12}}
Mean pairwise correlation of elicitation error vectors across the fixed paraphrases of the E17 battery versus across same-prompt resamples, per benchmark (source file transposed): paraphrase mixtures (AutoElicit-style) are as dependent as repeated draws (source \texttt{R12\_\allowbreak{}paraphrase\_\allowbreak{}errcorr.csv}).
\begin{rtab}{Table~R12}\setlength{\tabcolsep}{3.5pt}
\rtfitcol{%
\begin{tabular}{lrr}
\toprule
quantity & california & diabetes \\
\midrule
\# prompts (canonical $+$ five paraphrases) & 6 & 6 \\
error corr.\ across paraphrases & 0.993 & 0.971 \\
error corr.\ within prompt (resamples) & 0.986 & 0.987 \\
\bottomrule
\end{tabular}}
\end{rtab}

\subsection*{Table~R13: Predictive coverage of nominal-$90\%$ intervals at $n{=}20$, by method and benchmark}{\makeatletter\def\@currentlabel{R13}\makeatother\label{tab:R13}}
Empirical coverage of the nominal-$90\%$ predictive intervals at $n{=}20$ for every method (rows) on each Gaussian benchmark (columns; last column = mean over benchmarks, the quantity quoted in the text); the source file is transposed here (source \texttt{R13\_\allowbreak{}coverage\_\allowbreak{}by\_\allowbreak{}benchmark.csv}).
\begin{rtab}{Table~R13}\setlength{\tabcolsep}{3.5pt}
\rtfitcol{%
\begin{tabular}{lrrrrr}
\toprule
method & calif\_real & california & diabetes & nyc2026 & mean \\
\midrule
cv\_ridge & 0.905 & 0.862 & 0.876 & 0.852 & 0.874 \\
eb\_zero\_iso & 0.886 & 0.849 & 0.881 & 0.859 & 0.869 \\
eb\_zero\_pc & 0.867 & 0.813 & 0.828 & 0.842 & 0.838 \\
flat & 0.911 & 0.877 & 0.879 & 0.872 & 0.885 \\
llm\_cal\_iso & 0.876 & 0.819 & 0.874 & 0.859 & 0.857 \\
llm\_cal\_pc & 0.867 & 0.803 & 0.825 & 0.845 & 0.835 \\
llm\_fixed & 0.907 & 0.858 & 0.889 & 0.853 & 0.877 \\
llm\_hier & 0.863 & 0.788 & 0.837 & 0.852 & 0.835 \\
llm\_logpool & 0.859 & 0.785 & 0.826 & 0.831 & 0.825 \\
llm\_moment & 0.902 & 0.849 & 0.889 & 0.852 & 0.873 \\
llm\_robust\_mix & 0.876 & 0.820 & 0.874 & 0.859 & 0.857 \\
weak\_zero\_iso & 0.911 & 0.877 & 0.892 & 0.874 & 0.889 \\
\bottomrule
\end{tabular}}
\end{rtab}

\subsection*{Table~R14: Robust mixture: retention of the aligned-side gain}{\makeatletter\def\@currentlabel{R14}\makeatother\label{tab:R14}}
At $n{=}20$ on the aligned Gaussian benchmarks: RMSE gain over the weak prior of the best LLM-using method, gain of the robust mixture, and their ratio (point estimates; the bootstrap intervals quoted in the text are computed on the same splits) (source \texttt{R14\_\allowbreak{}mix\_\allowbreak{}retention.csv}).
\begin{rtab}{Table~R14}\setlength{\tabcolsep}{3.5pt}
\rtfitcol{%
\begin{tabular}{lrrr}
\toprule
benchmark & best LLM gain & mixture gain & retention \\
\midrule
california & 0.023 & 0.006 & 0.265 \\
diabetes & 0.138 & 0.122 & 0.880 \\
nyc2026 & 0.114 & 0.105 & 0.922 \\
\bottomrule
\end{tabular}}
\end{rtab}

\subsection*{Table~\ref{tab:R16}: Sign surgery on the Diabetes prior}{\makeatletter\def\@currentlabel{R16}\makeatother\label{tab:R16}}
Flipping the sign of the $k$ largest-$|\beta^\star_j|$ coordinates of the accurate Diabetes prior: displacement, gate trust rate and mean $n{=}20$ held-out RMSE of the weak, fixed (isotropic scale $0.3$), calibrated and robust-mixture priors, plus the mean calibrated scale and posterior LLM-component weight, aggregated over the 50 paired splits per $k$ (per-split source \texttt{R16\_\allowbreak{}signsurgery.csv}).
\begin{rtab}{Table~\ref{tab:R16}}\setlength{\tabcolsep}{3.5pt}
\rtfitcol{%
\begin{tabular}{lrrrrr}
\toprule
metric & $k{=}0$ & $k{=}1$ & $k{=}2$ & $k{=}3$ & $k{=}4$ \\
\midrule
$\delta$ & 0.760 & 0.644 & 0.954 & 1.166 & 1.247 \\
gate trust rate & 0.840 & 0.760 & 0.260 & 0.040 & 0.060 \\
RMSE weak & 0.989 & 0.989 & 0.989 & 0.989 & 0.989 \\
RMSE fixed & 0.861 & 0.860 & 0.892 & 0.907 & 0.898 \\
RMSE calibrated & 0.847 & 0.844 & 0.904 & 0.915 & 0.905 \\
RMSE mixture & 0.847 & 0.844 & 0.904 & 0.915 & 0.906 \\
mean $\tau^{\star2}$ & 0.037 & 0.029 & 0.036 & 0.104 & 0.098 \\
mean $\bar w_{\mathrm{LLM}}$ & 0.999 & 0.999 & 0.995 & 0.977 & 0.981 \\
\bottomrule
\end{tabular}}
\end{rtab}

\subsection*{Table~\ref{tab:R17}a: Newer flash-generation replication: displacement by model}{\makeatletter\def\@currentlabel{R17}\makeatother\label{tab:R17}}
Displacement $\delta=\|\bm m_{\mathrm{LLM}}-\bm\beta^\star\|$, fraction of coefficient signs agreeing with $\bm\beta^\star$, Pearson correlation with $\bm\beta^\star$, and $\|\bm\beta^\star\|$ for the two primary elicitors and the newer \texttt{gemini-3-flash-preview} model on four benchmarks (source \texttt{R17\_\allowbreak{}frontier\_\allowbreak{}displacement.csv}).
\begin{rtab}{Table~\ref{tab:R17}a}\setlength{\tabcolsep}{3.5pt}
\rtfitcol{%
\begin{tabular}{llrrrr}
\toprule
benchmark & model & $\delta$ & sign agr. & Pearson $r$ & $\|\bm\beta^\star\|$ \\
\midrule
california & claude-opus-4-7 & 1.095 & 0.750 & 0.739 & 1.490 \\
california & gemini-3.1-flash-lite & 1.099 & 0.750 & 0.780 & 1.490 \\
california & gemini-3-flash-preview & 0.666 & 0.875 & 0.956 & 1.490 \\
\midrule
diabetes & claude-opus-4-7 & 0.760 & 0.600 & 0.436 & 0.851 \\
diabetes & gemini-3.1-flash-lite & 0.760 & 0.600 & 0.484 & 0.851 \\
diabetes & gemini-3-flash-preview & 0.650 & 0.600 & 0.609 & 0.851 \\
\midrule
bcancer & claude-opus-4-7 & 2.696 & 0.700 & 0.521 & 3.584 \\
bcancer & gemini-3.1-flash-lite & 3.112 & 0.800 & 0.569 & 3.584 \\
bcancer & gemini-3-flash-preview & 3.187 & 0.800 & 0.627 & 3.584 \\
\midrule
logistic & claude-opus-4-7 & 1.066 & 0.833 & 0.512 & 1.197 \\
logistic & gemini-3.1-flash-lite & 0.987 & 0.833 & 0.544 & 1.197 \\
logistic & gemini-3-flash-preview & 0.984 & 0.833 & 0.527 & 1.197 \\
\bottomrule
\end{tabular}}
\end{rtab}

\subsection*{Table~\ref{tab:R17}b: Cross-model error correlations including the newer flash generation}
Correlation of elicitation error vectors between model pairs per benchmark and pooled over benchmarks, with coordinate-bootstrap $95\%$ interval; the last column is the number of coordinates $p$ entering the correlation, not a $p$-value (source \texttt{R17\_\allowbreak{}frontier\_\allowbreak{}errcorr.csv}).
\begin{rtab}{Table~\ref{tab:R17}b}\setlength{\tabcolsep}{3.5pt}
\rtfitcol{%
\begin{tabular}{lrrrr}
\toprule
benchmark & $r$ & $95\%$ lo & $95\%$ hi & \#coord. \\
\midrule
\multicolumn{5}{l}{\textit{pair: claude-opus-4-7 $\mid$ gemini-3-flash-preview}} \\
california & 0.960 & 0.898 & 0.997 & 8 \\
diabetes & 0.910 & 0.631 & 0.982 & 10 \\
bcancer & 0.982 & 0.931 & 0.999 & 10 \\
logistic & 0.923 & 0.848 & 0.998 & 6 \\
POOLED & 0.957 & 0.912 & 0.978 & 34 \\
\midrule
\multicolumn{5}{l}{\textit{pair: gemini-3.1-flash-lite $\mid$ gemini-3-flash-preview}} \\
california & 0.960 & 0.872 & 0.999 & 8 \\
diabetes & 0.976 & 0.851 & 0.997 & 10 \\
bcancer & 0.998 & 0.993 & 1.000 & 10 \\
logistic & 0.997 & 0.986 & 1.000 & 6 \\
POOLED & 0.986 & 0.966 & 0.995 & 34 \\
\midrule
\multicolumn{5}{l}{\textit{pair: claude-opus-4-7 $\mid$ gemini-3.1-flash-lite}} \\
california & 0.990 & 0.950 & 1.000 & 8 \\
diabetes & 0.932 & 0.663 & 0.987 & 10 \\
bcancer & 0.985 & 0.938 & 0.999 & 10 \\
logistic & 0.892 & 0.772 & 0.995 & 6 \\
POOLED & 0.970 & 0.937 & 0.987 & 34 \\
\bottomrule
\end{tabular}}
\end{rtab}

\subsection*{Table~\ref{tab:R18}a: Error correlations after projecting out the marginal-vs-partial direction}{\makeatletter\def\@currentlabel{R18}\makeatother\label{tab:R18}}
Cross-model error correlation per benchmark and model pair before and after projecting out the shared marginal-vs-partial discrepancy direction: correlations drop on the collinear real designs but survive on the randomized stress test (source \texttt{R18\_\allowbreak{}errcorr\_\allowbreak{}projected.csv}).
\begin{rtab}{Table~\ref{tab:R18}a}\setlength{\tabcolsep}{3.5pt}
\rtfitcol{%
\begin{tabular}{llrr}
\toprule
benchmark & pair & $r$ raw & $r$ proj. \\
\midrule
california & claude $\mid$ gemini-3.1-flash-lite & 0.990 & 0.775 \\
california & claude $\mid$ gemini-3-flash-preview & 0.960 & 0.230 \\
california & gemini $\mid$ gemini-3-flash-preview & 0.960 & 0.086 \\
\midrule
diabetes & claude $\mid$ gemini-3.1-flash-lite & 0.932 & 0.648 \\
diabetes & claude $\mid$ gemini-3-flash-preview & 0.910 & 0.613 \\
diabetes & gemini $\mid$ gemini-3-flash-preview & 0.976 & 0.903 \\
\midrule
bcancer & claude $\mid$ gemini-3.1-flash-lite & 0.985 & 0.749 \\
bcancer & claude $\mid$ gemini-3-flash-preview & 0.982 & 0.774 \\
bcancer & gemini $\mid$ gemini-3-flash-preview & 0.998 & 0.988 \\
\midrule
logistic & claude $\mid$ gemini-3.1-flash-lite & 0.892 & 0.962 \\
logistic & claude $\mid$ gemini-3-flash-preview & 0.923 & 0.984 \\
logistic & gemini $\mid$ gemini-3-flash-preview & 0.997 & 0.994 \\
\bottomrule
\end{tabular}}
\end{rtab}

\subsection*{Table~\ref{tab:R18}b: Estimand decomposition of the elicitation displacement}
Per benchmark and model: squared displacement $\delta^2$, fraction of $\delta^2$ lying along the marginal$\to$partial discrepancy direction $\bm g$, correlation of the error vector with $\bm g$, and correlations of the elicited mean with the per-feature simple-regression (marginal) slopes and with the OLS $\bm\beta^\star$ it is scored against (source \texttt{R18\_\allowbreak{}estimand\_\allowbreak{}decomp.csv}).
\begin{rtab}{Table~\ref{tab:R18}b}\setlength{\tabcolsep}{3.5pt}
\rtfitcol{%
\begin{tabular}{llrrrrr}
\toprule
benchmark & model & $\delta^2$ & frac.\ $\bm g$ & $r(\bm e,\bm g)$ & $r(\bm\mu,\mathrm{marg})$ & $r(\bm\mu,\bm\beta^\star)$ \\
\midrule
california & claude-opus-4-7 & 1.199 & 0.945 & 0.976 & 0.955 & 0.739 \\
california & gemini-3.1-flash-lite & 1.208 & 0.971 & 0.986 & 0.976 & 0.780 \\
california & gemini-3-flash-preview & 0.444 & 0.948 & 0.969 & 0.868 & 0.956 \\
\midrule
diabetes & claude-opus-4-7 & 0.577 & 0.737 & 0.910 & 0.972 & 0.436 \\
diabetes & gemini-3.1-flash-lite & 0.578 & 0.761 & 0.915 & 0.929 & 0.484 \\
diabetes & gemini-3-flash-preview & 0.422 & 0.655 & 0.883 & 0.886 & 0.609 \\
\midrule
bcancer & claude-opus-4-7 & 7.268 & 0.924 & 0.980 & 0.824 & 0.521 \\
bcancer & gemini-3.1-flash-lite & 9.683 & 0.861 & 0.976 & 0.923 & 0.569 \\
bcancer & gemini-3-flash-preview & 10.159 & 0.837 & 0.966 & 0.778 & 0.627 \\
\midrule
logistic & claude-opus-4-7 & 1.136 & 0.330 & 0.624 & 0.473 & 0.512 \\
logistic & gemini-3.1-flash-lite & 0.973 & 0.803 & 0.898 & 0.478 & 0.544 \\
logistic & gemini-3-flash-preview & 0.968 & 0.734 & 0.869 & 0.474 & 0.527 \\
\bottomrule
\end{tabular}}
\end{rtab}

\subsection*{Table~\ref{tab:R21}: Permuted-semantics probe battery: pre-specified summary}{\makeatletter\def\@currentlabel{R21}\makeatother\label{tab:R21}}
Summary statistics of the powered second-stage probe (\texttt{gemini-3.1-flash-lite}): number of derangements with $D>0$ out of $R$, sign-test $p$-value, Spearman $\rho$ of column-tracking against the number of permuted descriptions ($0$, the partial-permutation arm's $2$, $4$, $6$, and $8$) with its randomization $p$-value, and the pre-specified verdict (source \texttt{R21\_\allowbreak{}probe\_\allowbreak{}battery.json}).
\begin{rtab}{Table~\ref{tab:R21}}\setlength{\tabcolsep}{3.5pt}
\rtfitcol{%
\begin{tabular}{lp{1.6in}}
\toprule
key & value \\
\midrule
n\_pos & 20 \\
R & 20 \\
p\_sign & $9.5\times10^{-7}$ \\
rho & $-0.548$ \\
p\_dose & 0.002 \\
verdict & SEMANTIC (registered) \\
\bottomrule
\end{tabular}}
\end{rtab}

\subsection*{Table~R22: Partial-effect (ceteris-paribus) re-elicitation vs.\ the original semantic prompt}{\makeatletter\def\@currentlabel{R22}\makeatother\label{tab:R22}}
Re-eliciting with the full feature-correlation matrix disclosed and an explicit ceteris-paribus instruction (\texttt{PARTIAL}) versus the original semantic prompt: displacement, correlation of the elicited mean with $\bm\beta^\star$ and with the marginal slopes, fraction of $\delta^2$ along the discrepancy direction $\bm g$, and sign agreement; prompting does not remove the estimand effect (source \texttt{R22\_\allowbreak{}partial\_\allowbreak{}elicitation\_\allowbreak{}standardized.csv}).
\begin{rtab}{Table~R22}\setlength{\tabcolsep}{3.5pt}
\rtfitcol{%
\begin{tabular}{llrrrrr}
\toprule
dataset & prompt & $\delta$ & $r(\bm\mu,\bm\beta^\star)$ & $r(\bm\mu,\mathrm{marg})$ & frac.\ $\bm g$ & sign \\
\midrule
nyc2026 & orig SEM & 0.455 & 0.713 & 0.652 & 0.100 & 0.625 \\
nyc2026 & PARTIAL & 0.419 & 0.681 & 0.829 & 0.420 & 0.625 \\
\midrule
calif\_real & orig SEM & 0.971 & 0.786 & 0.977 & 0.978 & 0.750 \\
calif\_real & PARTIAL & 1.037 & 0.733 & 0.970 & 0.961 & 0.750 \\
\bottomrule
\end{tabular}}
\end{rtab}

\subsection*{Table~R26: Calibration of elicited scales $\sigma_j$}{\makeatletter\def\@currentlabel{R26}\makeatother\label{tab:R26}}
For each benchmark and elicitor (\texttt{gemini} = \texttt{gemini-3.1-flash-lite}, \texttt{gemini-3f} = \texttt{gemini-3-flash-preview}): median standardized error $|\mu_j-\beta^\star_j|/\sigma_j$ of the elicited coordinates, and the fraction of coordinates whose elicited $95\%$ and $90\%$ intervals cover $\beta^\star_j$, over the $p$ coefficients (last column) (source \texttt{R26\_\allowbreak{}width\_\allowbreak{}calibration.csv}).
\begin{rtab}{Table~R26}\setlength{\tabcolsep}{3.5pt}
\rtfitcol{%
\begin{tabular}{llrrrr}
\toprule
benchmark & model & median $|z|$ & $95\%$ cov. & $90\%$ cov. & $p$ (coord.) \\
\midrule
california & claude & 0.950 & 0.875 & 0.750 & 8 \\
california & gemini & 2.820 & 0.500 & 0.500 & 8 \\
california & gemini-3f & 0.720 & 0.750 & 0.750 & 8 \\
\midrule
diabetes & claude & 0.500 & 1.000 & 0.900 & 10 \\
diabetes & gemini & 0.790 & 0.900 & 0.800 & 10 \\
diabetes & gemini-3f & 0.640 & 0.900 & 0.900 & 10 \\
\midrule
bcancer & claude & 2.000 & 0.400 & 0.400 & 10 \\
bcancer & gemini & 4.590 & 0.100 & 0.100 & 10 \\
bcancer & gemini-3f & 2.640 & 0.200 & 0.200 & 10 \\
\midrule
logistic & claude & 1.380 & 0.667 & 0.667 & 6 \\
logistic & gemini & 2.590 & 0.500 & 0.167 & 6 \\
logistic & gemini-3f & 2.930 & 0.333 & 0.333 & 6 \\
\bottomrule
\end{tabular}}
\end{rtab}

\subsection*{Table~R27: Random-direction null for the estimand decomposition}{\makeatletter\def\@currentlabel{R27}\makeatother\label{tab:R27}}
Observed fraction of $\delta^2$ along the marginal-vs-partial discrepancy direction, its mean under a random-direction null with the null $p$-value, and the fraction re-estimated under cross-fitting of the direction (one random split); the mechanism is claimed only where the cross-fit fraction is stable (California) (source \texttt{R27\_\allowbreak{}estimand\_\allowbreak{}null.csv}).
\begin{rtab}{Table~R27}\setlength{\tabcolsep}{3.5pt}
\rtfitcol{%
\begin{tabular}{lrrrr}
\toprule
benchmark & observed frac. & null mean & null $p$ & cross-fit frac. \\
\midrule
california & 0.971 & 0.125 & 0 & 0.973 \\
diabetes & 0.761 & 0.100 & $2.0\times10^{-4}$ & 0.029 \\
\bottomrule
\end{tabular}}
\end{rtab}

\restoreintextsep\renewcommand{\textfraction}{0.05}

\section{Tier-2 Batteries: Pre-Specified External Validation and Prompt Variants}\label{app:t2}

Every battery in this section uses the model \gemini, $K{=}3$
draws at $T{=}0.7$ ($K{=}1$ per derangement in the permuted battery),
the canonical prompt template (Appendix~\ref{app:prompt}) unless
stated, and the frozen evaluation engine of
\texttt{FROZEN\_PROTOCOL.md} (written 2026-08-03, before any
candidate was retrieved): $n\in\{20,50\}$, $40$
paired splits, $\sigma^2$ the full-data OLS
residual variance, methods weak / moment-matched / calibrated
isotropic / robust mixture / EB-iso / EB-pc / CV ridge / flat, the
deployed gate $\|\bm m-\widehat{\bm\beta}\|\le\|\widehat{\bm\beta}\|$,
and the policies always-weak, always-moment, always-mixture, gated.
The addendum \texttt{PREREGISTRATION\_T2.md} (2026-09-03, before the first call)
fixes the implementation decisions the frozen text leaves open and
the pre-stated readings quoted below; transcripts of every call ship
with the code.  Neither file was lodged with a third-party registry,
contrary to what the submitted version announced.

\paragraph{Frozen protocol: dataset selection.}  Sources in fixed
order (UCI regression index, then Socrata portals, then Hugging
Face), candidates sorted lexicographically by identifier, filters
F1--F6 with a weak-semantics quota (CSV without auth; $n\ge300$ after dropping missing rows; $5\le
p\le10$ numeric non-identifier features; documented target else the
last numeric column, $\ge20$ distinct values, log-transformed if
strictly positive with skew $>2$; not used elsewhere in the paper;
domain cap $2$; at least two weak-semantics domains by rule), accept
until six pass.  Implementation decisions and two corrections made
before any analysis are disclosed: a column counts as numeric when
$\ge95\%$ of its non-missing entries parse as numbers; the source's
documented missing symbol counts as missing ($-200$ in Air Quality,
stated only in its free text); the identifier regex
\texttt{/id|code|key|index|zip|fips/i} is applied literally as
written, which drops \texttt{holiday} from Bike Sharing, and UCI's
own ``Categorical'' type labels exclude the PT08 sensor channels of
Air Quality, whose rule-selected target then becomes absolute
humidity.  Filter F5 is applied as a name filter that also excludes
datasets bundled with scikit-learn, R or seaborn; it rejected five
candidates, listed in the attempt log.  The six selected problems (Table~\ref{tab:t2datasets})
are all weak-semantics by rule; the full attempt log ships with the
code.  Dataset creators: Forest Fires \citep{cortez2007}, Concrete
\citep{yeh1998}, Energy Efficiency \citep{tsanas2012}, Istanbul Stock
Exchange \citep{akbilgic2014}, Bike Sharing \citep{fanaee2014}, Air
Quality \citep{devito2008}; repositories: UCI \citep{uci2023}, OpenML
and OpenML-CTR23 \citep{vanschoren2014,fischer2023ctr23}, PMLB
\citep{olson2017pmlb}.

\begin{table*}[tbp]\centering\footnotesize
\caption{Rule-selected external datasets (elicitor \gemini). Top: the six frozen-protocol UCI datasets. Bottom: the pre-specified extension (OpenML-CTR23, then PMLB; $4\le p\le10$, $n\ge500$, readable feature names; excluded are the synthetic families flagged by name, Friedman/\texttt{fri\_c}, \texttt{pwLinear} and \texttt{chscase}, while the Bayesian-network-generated \texttt{BNG} sets and the simulated \texttt{pollen} passed the name filter). ex03 and ex04 duplicate fz02 and fz03 and are counted once, by their fz rows.}
\label{tab:t2datasets}
\resizebox{\linewidth}{!}{%
\begin{tabular}{lllrrll}\toprule
name & source name & identifier & $n$ & $p$ & target & category \\\midrule
fz01\_forestfires & Forest Fires & uci-162 & 517 & 10 & area & climate and environment \\
fz02\_concretecompress & Concrete Compressive Strength & uci-165 & 1030 & 8 & Concrete compressive strength & physics and chemistry \\
fz03\_energyefficiency & Energy Efficiency & uci-242 & 768 & 8 & Y1 & computer science \\
fz04\_istanbulstockexc & ISTANBUL STOCK EXCHANGE & uci-247 & 536 & 8 & EM & business \\
fz05\_bikesharing & Bike Sharing & uci-275 & 17379 & 5 & cnt & social science \\
fz06\_airquality & Air Quality & uci-360 & 827 & 7 & AH & computer science \\
\midrule
ex01\_airfoilselfnoise & airfoil\_self\_noise & openml-044957 & 1503 & 5 & sound\_pressure & OpenML-CTR23 \\
ex02\_auctionverificat & auction\_verification & openml-044958 & 2043 & 7 & verification.time & OpenML-CTR23 \\
ex03\_concretecompress & concrete\_compressive\_strength & openml-044959 & 1030 & 8 & strength & OpenML-CTR23 \\
ex04\_energyefficiency & energy\_efficiency & openml-044960 & 768 & 8 & heating\_load & OpenML-CTR23 \\
ex05\_qsarfishtoxicity & QSAR\_fish\_toxicity & openml-044970 & 908 & 6 & LC50 & OpenML-CTR23 \\
ex06\_kin8nm & kin8nm & openml-044980 & 8192 & 8 & y & OpenML-CTR23 \\
ex07\_healthinsurance & health\_insurance & openml-044993 & 22272 & 4 & whrswk & OpenML-CTR23 \\
ex08\_spacega & space\_ga & openml-045402 & 3107 & 5 & ln\_votes\_pop & OpenML-CTR23 \\
ex09\_1196bngpharynx & 1196\_BNG\_pharynx & pmlb-1196\_BNG\_pharynx & 1000000 & 10 & target & PMLB \\
ex10\_1199bngechomonth & 1199\_BNG\_echoMonths & pmlb-1199\_BNG\_echoMonths & 17496 & 8 & target & PMLB \\
ex11\_225puma8nh & 225\_puma8NH & pmlb-225\_puma8NH & 8192 & 8 & target & PMLB \\
ex12\_522pm10 & 522\_pm10 & pmlb-522\_pm10 & 500 & 7 & target & PMLB \\
ex13\_529pollen & 529\_pollen & pmlb-529\_pollen & 3848 & 4 & target & PMLB \\
ex14\_666rmftsaladata & 666\_rmftsa\_ladata & pmlb-666\_rmftsa\_ladata & 508 & 10 & target & PMLB \\
\bottomrule\end{tabular}}
\end{table*}

\paragraph{Outcomes against the pre-stated readings.}
\emph{Frozen verdict:} \textsc{partial}, with gated relative regret 2.4\% on
distrusted cells ($\le3\%$) and 3.2\% on trusted cells
($\le9\%$); one dataset (Air Quality) has a trusted cell where the
moment prior is more than $5\%$ worse than weak
(Table~\ref{tab:t2frozen}).  \emph{Extension verdict:}
\textsc{partial} (one violating dataset, space\_ga;
Table~\ref{tab:t2extension}); the pre-stated tally over the $18$
distinct external problems at $n{=}20$ is $6$ better than the
strongest classical prior with a CI excluding zero, $8$ within
$[-1.2,+3.7]\%$, $4$ worse by more than $5\%$.
\emph{Permuted descriptions} (Table~\ref{tab:t2perm}): the
calibrated prior meets the pre-stated $\le3\%$ median-harm criterion
on $25$ of the $26$ regression problems (exception: the OpenML copy
of Energy Efficiency, $+5.4\%$); the gate's trust rate is lower on permuted
than on true priors on $16/28$ problems, below the pre-stated
$80\%$.  But on $17/28$ problems the permuted prior is the true prior
(median correlation $\ge0.90$; column names anchor the elicitation),
so there is nothing to detect, and on the problems where the
permutation moved the prior the trust rate fell (Energy
$0.85\to0.12$, OpenML Energy $0.85\to0.31$, Stokes $1.00\to0.53$).
\emph{LLM paraphrases} (Table~\ref{tab:t2para}): errors are shared by
the pre-stated criterion (mean error correlation $0.92$--$0.98$,
lower bounds $>0.70$); the ensemble over all paraphrase draws does
not repair the center ($\delta$ ratio $\ge0.91$ on every problem, vs.\
the pre-stated $\le0.8$).  \emph{No-default anonymized prompt}
(Table~\ref{tab:t2nodefault}): sign agreement lies in $[0.35,0.65]$
on $71\%$ of problems and the correlation with the semantic prior is
below $0.3$ on $56\%$ of the $27$ problems where it is defined,
short of the pre-stated $80\%$; the residual
correlations are explained by a content-free positional template
(over all draws, $\mu_1>0$ in $100\%$, $\mu_2<0$ in $94\%$,
$\mu_3>0$ in $95\%$, $\mu_4<0$ in $83\%$, magnitudes decaying from
$0.20$), which coincides with the true sign pattern on datasets whose
first column is the dominant positive driver (California $0.63$,
Breast Cancer $0.66$).  Against the strongest classical prior the
forced prior is worse on $20$ of $26$ regression problems (median
$+1.2\%$).  \emph{Derivable problems} (Table~\ref{tab:t2derivable}):
the pre-stated ``derives'' reading (correlation $\ge0.9$ with the
analytic coefficients and full sign agreement on active coefficients
on at least two of three) is met on all three.  \emph{Newer flash
models:} the pre-specified battery of \geminew{} over the new
datasets was not run (no API calls were made after the review
cycle); only its canonical-benchmark results (Table~\ref{tab:R17})
exist.

\begin{table*}[tbp]\centering\footnotesize
\caption{Frozen-protocol external validation on the six rule-selected UCI datasets (\gemini, $K{=}3$): RMSE change (\%) of the moment-matched and calibrated LLM priors against the best zero-centered prior in the cell, the gate's trust rate, and the relative regret of the gated policy and of the robust mixture. Verdict: PARTIAL.}\label{tab:t2frozen}
\resizebox{\linewidth}{!}{%
\begin{tabular}{lrrrrrrrrrrr}\toprule
dataset & $n$ & best classical & moment vs best (\%) & 95\% CI & cal vs best (\%) & 95\% CI & moment vs weak (\%) & gate & trust rate & rel.\ regret (gate) & rel.\ regret (mix) \\\midrule
fz01\_forestfires & 20 & eb\_zero\_iso & +0.6 & [0.0, 1.2] & +0.8 & [0.3, 1.5] & $-$3.1 & distrust & 0.125 & 2.7\% & 0.3\% \\
fz01\_forestfires & 50 & eb\_zero\_iso & +2.3 & [0.9, 4.1] & +2.1 & [0.8, 3.8] & $-$1.8 & distrust & 0.000 & 2.1\% & 0.0\% \\
fz02\_concretecompress & 20 & eb\_zero\_iso & $-$4.7 & [$-$7.9, $-$1.6] & $-$8.5 & [$-$11.0, $-$5.9] & $-$12.2 & trust & 0.975 & 4.7\% & 0.0\% \\
fz02\_concretecompress & 50 & eb\_zero\_iso & $-$3.0 & [$-$4.0, $-$2.0] & $-$3.9 & [$-$5.6, $-$2.4] & $-$4.2 & trust & 1.000 & 0.9\% & 0.0\% \\
fz03\_energyefficiency & 20 & eb\_zero\_pc & $-$7.4 & [$-$10.5, $-$4.6] & $-$1.5 & [$-$4.1, 1.2] & $-$7.4 & trust & 0.850 & 1.5\% & 6.5\% \\
fz03\_energyefficiency & 50 & weak & $-$1.2 & [$-$2.3, $-$0.3] & $-$0.0 & [$-$0.6, 0.5] & $-$1.2 & trust & 0.975 & 0.1\% & 1.2\% \\
fz04\_istanbulstockexc & 20 & eb\_zero\_iso & $-$1.7 & [$-$3.2, $-$0.2] & $-$0.7 & [$-$1.7, 0.4] & $-$6.5 & trust & 0.700 & 3.0\% & 1.2\% \\
fz04\_istanbulstockexc & 50 & eb\_zero\_iso & $-$1.1 & [$-$1.8, $-$0.3] & $-$0.7 & [$-$1.0, $-$0.4] & $-$3.5 & trust & 0.800 & 1.3\% & 0.4\% \\
fz05\_bikesharing & 20 & eb\_zero\_iso & $-$2.8 & [$-$4.7, $-$1.2] & $-$1.6 & [$-$3.2, 0.3] & $-$4.6 & trust & 0.575 & 2.7\% & 1.4\% \\
fz05\_bikesharing & 50 & eb\_zero\_iso & $-$1.2 & [$-$2.1, $-$0.3] & $-$0.3 & [$-$1.0, 0.5] & $-$2.0 & trust & 0.575 & 1.1\% & 1.0\% \\
fz06\_airquality & 20 & eb\_zero\_pc & +17.5 & [10.4, 25.0] & +4.3 & [1.4, 7.1] & +9.0 & trust & 1.000 & 12.5\% & 0.0\% \\
fz06\_airquality & 50 & eb\_zero\_pc & +5.3 & [2.7, 8.0] & +1.3 & [0.4, 2.3] & +2.8 & trust & 1.000 & 3.9\% & 0.0\% \\
\bottomrule\end{tabular}}
\end{table*}

\begin{table*}[tbp]\centering\footnotesize
\caption{Extension battery (OpenML-CTR23 and PMLB; \gemini, $K{=}3$), same columns as Table~\ref{tab:t2frozen}. Verdict: PARTIAL.}\label{tab:t2extension}
\resizebox{\linewidth}{!}{%
\begin{tabular}{lrrrrrrrrrrr}\toprule
dataset & $n$ & best classical & moment vs best (\%) & 95\% CI & cal vs best (\%) & 95\% CI & moment vs weak (\%) & gate & trust rate & rel.\ regret (gate) & rel.\ regret (mix) \\\midrule
ex01\_airfoilselfnoise & 20 & eb\_zero\_iso & +2.8 & [1.0, 4.5] & +1.4 & [$-$0.4, 3.6] & $-$0.7 & distrust & 0.125 & 2.3\% & 0.0\% \\
ex01\_airfoilselfnoise & 50 & eb\_zero\_iso & +0.4 & [$-$0.4, 1.3] & +0.2 & [$-$0.2, 0.5] & $-$0.2 & distrust & 0.025 & 0.3\% & 0.0\% \\
ex02\_auctionverificat & 20 & eb\_zero\_iso & $-$2.0 & [$-$2.9, $-$1.1] & +0.7 & [$-$0.1, 1.5] & $-$7.3 & distrust & 0.425 & 4.7\% & 2.8\% \\
ex02\_auctionverificat & 50 & eb\_zero\_pc & +0.8 & [0.3, 1.2] & +1.8 & [1.2, 2.3] & $-$1.4 & distrust & 0.300 & 0.9\% & 1.0\% \\
ex03\_concretecompress & 20 & eb\_zero\_iso & $-$3.7 & [$-$7.2, $-$0.2] & $-$6.0 & [$-$8.4, $-$3.5] & $-$11.3 & trust & 0.950 & 3.5\% & 0.0\% \\
ex03\_concretecompress & 50 & eb\_zero\_iso & $-$2.8 & [$-$3.8, $-$1.8] & $-$2.9 & [$-$4.3, $-$1.7] & $-$4.0 & trust & 1.000 & 0.2\% & 0.0\% \\
ex04\_energyefficiency & 20 & eb\_zero\_pc & +0.3 & [$-$3.7, 3.7] & +3.0 & [0.2, 5.9] & +0.2 & trust & 0.850 & 0.3\% & 3.0\% \\
ex04\_energyefficiency & 50 & weak & +1.6 & [$-$0.4, 3.1] & +3.6 & [2.3, 4.8] & +1.6 & trust & 0.775 & 0.7\% & 3.6\% \\
ex05\_qsarfishtoxicity & 20 & eb\_zero\_iso & +2.0 & [0.6, 3.4] & +0.1 & [$-$0.6, 1.0] & $-$4.1 & distrust & 0.400 & 4.2\% & 0.0\% \\
ex05\_qsarfishtoxicity & 50 & eb\_zero\_iso & +0.7 & [0.3, 1.0] & +0.3 & [0.0, 0.6] & $-$0.6 & distrust & 0.250 & 0.8\% & 0.0\% \\
ex06\_kin8nm & 20 & eb\_zero\_iso & $-$0.5 & [$-$1.4, 0.1] & +0.0 & [0.0, 0.0] & $-$8.9 & trust & 1.000 & 0.0\% & 0.6\% \\
ex06\_kin8nm & 50 & eb\_zero\_iso & $-$0.3 & [$-$0.5, $-$0.1] & +0.0 & [0.0, 0.0] & $-$1.1 & trust & 1.000 & 0.0\% & 0.3\% \\
ex07\_healthinsurance & 20 & eb\_zero\_iso & +1.3 & [$-$0.2, 2.7] & +1.7 & [0.3, 3.2] & $-$2.4 & distrust & 0.250 & 2.8\% & 0.5\% \\
ex07\_healthinsurance & 50 & weak & +0.2 & [$-$0.8, 1.1] & $-$0.2 & [$-$0.8, 0.4] & +0.2 & distrust & 0.250 & 0.8\% & 0.0\% \\
ex08\_spacega & 20 & flat & +8.8 & [5.0, 12.5] & +4.7 & [1.5, 7.7] & +3.7 & trust & 0.675 & 3.9\% & 0.0\% \\
ex08\_spacega & 50 & flat & +15.0 & [12.8, 17.3] & +2.5 & [0.5, 4.7] & +8.4 & trust & 0.875 & 11.2\% & 0.0\% \\
ex09\_1196bngpharynx & 20 & eb\_zero\_pc & $-$2.9 & [$-$5.1, $-$0.8] & +4.4 & [2.0, 6.6] & $-$12.6 & distrust & 0.025 & 13.9\% & 7.5\% \\
ex09\_1196bngpharynx & 50 & eb\_zero\_pc & $-$0.0 & [$-$0.6, 0.6] & +2.1 & [1.4, 3.0] & $-$3.6 & distrust & 0.000 & 3.7\% & 2.2\% \\
ex10\_1199bngechomonth & 20 & eb\_zero\_iso & +14.1 & [11.9, 16.3] & +8.5 & [6.6, 10.6] & +1.3 & distrust & 0.000 & 3.0\% & 0.0\% \\
ex10\_1199bngechomonth & 50 & eb\_zero\_pc & +4.2 & [3.1, 5.3] & +2.4 & [1.7, 3.1] & +1.2 & distrust & 0.000 & 0.4\% & 0.0\% \\
ex11\_225puma8nh & 20 & eb\_zero\_pc & +3.7 & [1.7, 5.5] & +2.8 & [0.7, 4.7] & $-$6.1 & distrust & 0.250 & 5.5\% & 0.0\% \\
ex11\_225puma8nh & 50 & eb\_zero\_pc & +2.3 & [1.8, 2.9] & +2.2 & [1.5, 3.0] & $-$1.0 & distrust & 0.225 & 1.1\% & 0.0\% \\
ex12\_522pm10 & 20 & eb\_zero\_iso & $-$1.2 & [$-$2.4, 0.1] & $-$5.8 & [$-$6.6, $-$4.9] & $-$12.6 & trust & 0.625 & 10.6\% & 0.0\% \\
ex12\_522pm10 & 50 & eb\_zero\_pc & $-$1.3 & [$-$2.1, $-$0.4] & $-$4.6 & [$-$5.8, $-$3.4] & $-$3.7 & trust & 0.825 & 4.7\% & 0.0\% \\
ex13\_529pollen & 20 & eb\_zero\_pc & +57.4 & [52.4, 62.1] & +1.9 & [1.0, 2.8] & +55.4 & distrust & 0.000 & 0.0\% & 0.2\% \\
ex13\_529pollen & 50 & eb\_zero\_pc & +35.3 & [32.3, 38.3] & +0.3 & [$-$0.0, 0.7] & +35.0 & distrust & 0.000 & 0.0\% & 0.0\% \\
ex14\_666rmftsaladata & 20 & eb\_zero\_pc & +1.1 & [$-$2.5, 4.4] & +2.6 & [$-$0.2, 5.1] & $-$8.9 & trust & 0.625 & 6.7\% & 1.5\% \\
ex14\_666rmftsaladata & 50 & eb\_zero\_pc & +2.5 & [1.5, 3.4] & +3.6 & [2.9, 4.3] & $-$2.2 & trust & 0.825 & 1.2\% & 1.1\% \\
\bottomrule\end{tabular}}
\end{table*}

\begin{table*}[tbp]\centering\footnotesize
\caption{Permuted-description priors on all 28 problems (\gemini, $K{=}1$ per derangement): displacement of the true and permuted priors, the sign test $D=r_{\mathrm{perm}}-r_{\mathrm{id}}$ (Table~\ref{tab:R21}), harm of the calibrated prior built from a permuted mean at $n{=}20$ (\% vs.\ weak), and the gate's trust rate on true vs.\ permuted priors.}\label{tab:t2perm}
\resizebox{\linewidth}{!}{%
\begin{tabular}{lrrrrrrrrrrr}\toprule
benchmark & task & $\delta$ true & $\delta$ perm (med.) & frac $D>0$ & $p$ (semantic) & cal true vs weak (\%) & cal perm vs weak med.\ (\%) & max (\%) & frac harm $>5\%$ & gate trust (true) & gate trust (perm) \\\midrule
bcancer & logistic & 3.112 & 3.076 & 0.000 & 1.000 & 0.953 & 1.091 & 1.885 & 0.000 & 1.000 & 1.000 \\
california & regression & 1.099 & 1.111 & 0.000 & 1.000 & $-$0.053 & $-$0.043 & 6.202 & 0.050 & 1.000 & 0.993 \\
diabetes & regression & 0.760 & 0.723 & 0.250 & 0.994 & $-$14.006 & $-$14.311 & $-$6.380 & 0.000 & 0.800 & 0.753 \\
dv\_costfn & regression & 0.466 & 0.546 & 0.350 & 0.942 & $-$2.567 & $-$2.090 & 0.218 & 0.000 & 0.950 & 0.706 \\
dv\_pipeflow & regression & 0.439 & 0.496 & 0.150 & 1.000 & $-$10.189 & $-$9.186 & 0.484 & 0.000 & 1.000 & 0.685 \\
dv\_stokes & regression & 0.318 & 0.723 & 0.000 & 1.000 & $-$6.169 & $-$1.414 & 0.060 & 0.000 & 1.000 & 0.528 \\
ex01\_airfoilselfnoise & regression & 1.051 & 1.051 & 0.000 & 1.000 & $-$1.962 & $-$1.962 & $-$1.022 & 0.000 & 0.125 & 0.159 \\
ex02\_auctionverificat & regression & 0.665 & 0.685 & 0.150 & 1.000 & $-$4.733 & $-$4.364 & $-$2.596 & 0.000 & 0.425 & 0.320 \\
ex03\_concretecompress & regression & 0.591 & 0.583 & 0.000 & 1.000 & $-$13.385 & $-$14.189 & $-$10.568 & 0.000 & 0.950 & 0.954 \\
ex04\_energyefficiency & regression & 1.176 & 1.444 & 0.300 & 0.979 & 2.955 & 5.388 & 7.671 & 0.550 & 0.850 & 0.311 \\
ex05\_qsarfishtoxicity & regression & 0.675 & 0.458 & 0.700 & 0.058 & $-$5.795 & $-$6.600 & $-$0.090 & 0.000 & 0.400 & 0.559 \\
ex06\_kin8nm & regression & 0.646 & 0.646 & 0.000 & -- & $-$8.391 & $-$8.391 & $-$5.241 & 0.000 & 1.000 & 0.905 \\
ex07\_healthinsurance & regression & 0.545 & 0.594 & 0.000 & 1.000 & $-$1.951 & $-$1.636 & $-$1.486 & 0.000 & 0.250 & 0.244 \\
ex08\_spacega & regression & 6.004 & 6.007 & 0.000 & 1.000 & $-$0.313 & $-$0.318 & 0.583 & 0.000 & 0.675 & 0.695 \\
ex09\_1196bngpharynx & regression & 0.837 & 0.822 & 0.263 & 0.990 & $-$6.092 & $-$6.381 & $-$2.119 & 0.000 & 0.025 & 0.151 \\
ex10\_1199bngechomonth & regression & 1.275 & 1.120 & 0.000 & 1.000 & $-$3.672 & $-$4.892 & $-$3.264 & 0.000 & 0.000 & 0.004 \\
ex11\_225puma8nh & regression & 0.676 & 0.667 & 0.263 & 0.990 & $-$6.892 & $-$7.417 & $-$4.807 & 0.000 & 0.250 & 0.351 \\
ex12\_522pm10 & regression & 0.135 & 0.385 & 0.100 & 1.000 & $-$16.600 & $-$12.834 & $-$9.546 & 0.000 & 0.625 & 0.554 \\
ex13\_529pollen & regression & 2.522 & 2.475 & 0.222 & 0.980 & 0.570 & 0.553 & 0.659 & 0.000 & 0.000 & 0.000 \\
ex14\_666rmftsaladata & regression & 1.205 & 1.253 & 0.050 & 1.000 & $-$7.616 & $-$5.695 & $-$4.671 & 0.000 & 0.625 & 0.642 \\
fz01\_forestfires & regression & 0.425 & 0.524 & 0.150 & 1.000 & $-$2.828 & $-$2.165 & $-$1.431 & 0.000 & 0.125 & 0.065 \\
fz02\_concretecompress & regression & 0.602 & 0.586 & 0.000 & 1.000 & $-$15.721 & $-$15.681 & $-$12.502 & 0.000 & 0.975 & 0.953 \\
fz03\_energyefficiency & regression & 1.043 & 1.393 & 0.800 & 0.006 & $-$1.555 & 1.990 & 8.560 & 0.300 & 0.850 & 0.122 \\
fz04\_istanbulstockexc & regression & 0.574 & 0.593 & 0.500 & 0.588 & $-$5.577 & $-$3.725 & $-$0.837 & 0.000 & 0.700 & 0.684 \\
fz05\_bikesharing & regression & 0.376 & 0.380 & 0.000 & 1.000 & $-$3.378 & $-$2.710 & $-$0.989 & 0.000 & 0.575 & 0.554 \\
fz06\_airquality & regression & 1.054 & 1.058 & 0.100 & 1.000 & $-$3.248 & $-$3.111 & 1.304 & 0.000 & 1.000 & 0.951 \\
logistic & logistic & 0.987 & 0.952 & 0.000 & 1.000 & $-$3.952 & $-$4.367 & $-$2.240 & 0.000 & 0.825 & 0.844 \\
nyc2026 & regression & 0.455 & 0.497 & 0.000 & 1.000 & $-$7.729 & $-$7.636 & $-$4.982 & 0.000 & 0.350 & 0.410 \\
\bottomrule\end{tabular}}
\end{table*}

\begin{table*}[tbp]\centering\footnotesize
\caption{Twenty LLM-written paraphrases of the prompt (\gemini, $K{=}3$ each): displacement of the canonical prior, of the paraphrases (min, max) and of the AutoElicit-style ensemble over all paraphrase draws; mean error correlation across paraphrases (2.5/97.5 percentiles); RMSE at $n{=}20$ vs.\ weak.}\label{tab:t2para}
\resizebox{\linewidth}{!}{%
\begin{tabular}{lrrrrrrrrrr}\toprule
benchmark & \#para & $\delta$ canonical & $\delta$ ensemble & $\delta$ min & $\delta$ max & error corr. & lo & hi & moment canon vs weak (\%) & moment ens.\ vs weak (\%) \\\midrule
california & 20 & 1.099 & 1.285 & 1.100 & 1.490 & 0.978 & 0.915 & 0.998 & $-$2.071 & $-$0.771 \\
diabetes & 20 & 0.760 & 0.697 & 0.670 & 0.851 & 0.944 & 0.816 & 0.995 & $-$15.899 & $-$8.353 \\
nyc2026 & 20 & 0.455 & 0.533 & 0.452 & 1.223 & 0.919 & 0.705 & 0.989 & $-$8.642 & $-$9.415 \\
\bottomrule\end{tabular}}
\end{table*}

\begin{table*}[tbp]\centering\footnotesize
\caption{Anonymized prompt with the zero-mean default forbidden (\gemini, $K{=}3$): displacement and sign agreement of the forced prior, its correlation with the semantic prior, and its change in loss against weak at $n{=}20$ (negative is better).}\label{tab:t2nodefault}
\resizebox{\linewidth}{!}{%
\begin{tabular}{lrrrrrrrrrr}\toprule
benchmark & task & $\delta$ no-default & sign agr. & mean $|\mu|$ & $\delta$ anon & $\delta$ sem & corr(no-default, sem) & fixed vs weak (\%) & cal vs weak (\%) & gate trust \\\midrule
california & regression & 1.439 & 0.375 & 0.094 & 1.490 & 1.099 & 0.631 & 8.328 & 3.588 & 0.925 \\
diabetes & regression & 0.946 & 0.400 & 0.089 & 0.851 & 0.760 & 0.029 & $-$10.404 & $-$12.354 & 0.250 \\
bcancer & logistic & 3.511 & 0.600 & 0.053 & 3.584 & 3.112 & 0.662 & 32.845 & 1.147 & 1.000 \\
logistic & logistic & 1.238 & 0.667 & 0.139 & 1.197 & 0.987 & 0.051 & $-$2.400 & $-$1.311 & 0.425 \\
nyc2026 & regression & 0.613 & 0.750 & 0.110 & 0.679 & 0.455 & 0.403 & $-$7.458 & $-$5.690 & 0.550 \\
dv\_costfn & regression & 0.668 & 0.500 & 0.161 & 0.718 & 0.466 & 0.723 & $-$2.217 & $-$1.708 & 0.975 \\
dv\_pipeflow & regression & 0.602 & 0.875 & 0.098 & 0.839 & 0.439 & 0.750 & $-$8.096 & $-$6.362 & 1.000 \\
dv\_stokes & regression & 0.714 & 0.667 & 0.068 & 0.696 & 0.318 & 0.505 & $-$5.426 & $-$2.472 & 1.000 \\
ex01\_airfoilselfnoise & regression & 1.102 & 0.600 & 0.147 & 0.916 & 1.051 & $-$0.584 & $-$3.130 & $-$2.107 & 0.000 \\
ex02\_auctionverificat & regression & 0.535 & 0.571 & 0.106 & 0.609 & 0.665 & 0.042 & $-$6.904 & $-$6.250 & 0.850 \\
ex03\_concretecompress & regression & 0.999 & 0.625 & 0.058 & 1.073 & 0.591 & 0.542 & $-$12.385 & $-$9.386 & 0.900 \\
ex04\_energyefficiency & regression & 1.323 & 0.375 & 0.087 & 1.165 & 1.176 & 0.260 & 4.590 & 6.560 & 0.050 \\
ex05\_qsarfishtoxicity & regression & 0.738 & 0.333 & 0.085 & 0.575 & 0.675 & $-$0.147 & $-$4.918 & $-$4.840 & 0.050 \\
ex06\_kin8nm & regression & 0.768 & 0.500 & 0.124 & 0.646 & 0.646 & -- & $-$5.654 & $-$6.479 & 0.075 \\
ex07\_healthinsurance & regression & 0.666 & 0.500 & 0.133 & 0.500 & 0.545 & 0.851 & $-$2.667 & $-$0.893 & 0.025 \\
ex08\_spacega & regression & 6.259 & 0.600 & 0.150 & 6.095 & 6.004 & $-$0.924 & 4.281 & 1.023 & 0.075 \\
ex09\_1196bngpharynx & regression & 0.766 & 0.600 & 0.121 & 0.609 & 0.837 & 0.386 & $-$6.218 & $-$6.957 & 0.075 \\
ex10\_1199bngechomonth & regression & 0.739 & 0.500 & 0.067 & 0.549 & 1.275 & 0.568 & $-$10.773 & $-$8.393 & 0.075 \\
ex11\_225puma8nh & regression & 0.769 & 0.500 & 0.123 & 0.633 & 0.676 & $-$0.012 & $-$7.478 & $-$6.794 & 0.150 \\
ex12\_522pm10 & regression & 0.486 & 0.429 & 0.124 & 0.407 & 0.135 & 0.153 & $-$11.339 & $-$11.000 & 0.275 \\
ex13\_529pollen & regression & 2.356 & 0.250 & 0.158 & 2.102 & 2.522 & $-$0.272 & 40.187 & 0.240 & 0.000 \\
ex14\_666rmftsaladata & regression & 1.052 & 0.500 & 0.097 & 1.256 & 1.205 & $-$0.006 & $-$8.915 & $-$7.880 & 0.850 \\
fz01\_forestfires & regression & 0.365 & 0.600 & 0.123 & 0.164 & 0.425 & 0.360 & $-$2.914 & $-$2.946 & 0.400 \\
fz02\_concretecompress & regression & 1.085 & 0.625 & 0.123 & 1.102 & 0.602 & 0.393 & $-$12.154 & $-$8.334 & 0.650 \\
fz03\_energyefficiency & regression & 1.325 & 0.250 & 0.043 & 1.165 & 1.043 & $-$0.797 & 4.706 & 4.132 & 0.025 \\
fz04\_istanbulstockexc & regression & 0.740 & 0.375 & 0.085 & 0.647 & 0.574 & $-$0.486 & $-$2.080 & $-$3.859 & 0.225 \\
fz05\_bikesharing & regression & 0.474 & 0.800 & 0.139 & 0.429 & 0.376 & $-$0.395 & $-$0.664 & $-$0.297 & 0.550 \\
fz06\_airquality & regression & 1.982 & 0.429 & 0.106 & 1.922 & 1.054 & $-$0.199 & 75.735 & 0.209 & 0.100 \\
\bottomrule\end{tabular}}
\end{table*}

\begin{table*}[tbp]\centering\footnotesize
\caption{Derivable problems (\gemini, $K{=}3$): how close the elicited center comes to the coefficients implied by the stated physics or economics; nuisance features have true coefficient zero.}\label{tab:t2derivable}
\resizebox{\linewidth}{!}{%
\begin{tabular}{lrrrrrrrr}\toprule
problem & $\delta$ vs OLS & $\delta$ vs analytic & corr vs analytic & sign agr.\ (active) & mean $|\mu|$ active & mean $|\mu|$ nuisance & mean elicited scale $\sigma_j$ & $\delta$ anon \\\midrule
dv\_pipeflow & 0.439 & 0.531 & 0.941 & 1.000 & 0.457 & 0.000 & 0.135 & 0.839 \\
dv\_stokes & 0.318 & 0.366 & 0.989 & 1.000 & 0.609 & 0.000 & 0.168 & 0.696 \\
dv\_costfn & 0.466 & 0.577 & 0.926 & 1.000 & 0.618 & 0.000 & 0.147 & 0.718 \\
\bottomrule\end{tabular}}
\end{table*}

\section{Method Glossary and Elicitation Configuration}\label{app:gloss}

\paragraph{Method glossary.}  \texttt{llm\_sem}: the single Gaussian
obtained from the $K{=}3$ samples by the law of total variance (mean
of the $\mu_j$; variance = mean elicited $\sigma_j^2$ plus
between-draw variance, diagonal).  \texttt{llm\_moment}: moment match
of the $K$-component mixture (Eq.~\ref{eq:mm}); for diagonal
components the two summaries differ only in the off-diagonal
between-draw covariances and are
numerically near-identical on every benchmark where both run
(California $n{=}20$: $0.6905$ vs.\ $0.6910$; Diabetes $0.8458$ vs.\
$0.8455$; NYC $1.0113$ vs.\ $1.0110$).

\emph{Which draws enter.}  \texttt{llm\_fixed} $=$ \texttt{llm\_sem}
uses the primary vendor's $K{=}3$ draws (1v).  The factorial's
aggregators \texttt{llm\_moment}, \texttt{llm\_logpool} and
\texttt{llm\_hier} pool every model with recorded draws (2v: \claude{}
and \gemini, $K{=}6$), which is why Diabetes shows $0.846$ for the
former and $0.834$ for the latter.  The T1 tables label aggregates 1v
or 2v; NYC has one vendor only.
\texttt{llm\_logpool}: unit-weight log pool (Eq.~\ref{eq:lp});
\texttt{llm\_hier}: random-effects pool (Appendix~\ref{app:crossmodel});
\texttt{llm\_cal\_iso}/\texttt{llm\_cal\_pc}: translated evidence/ARD
calibration around $\mllm$; \texttt{llm\_robust\_mix}: two-component
mixture with evidence weights; zero-centered baselines as named.
\texttt{cv\_ridge} specification: $k=\min(5,\lfloor n/4\rfloor)$
folds ($5$ at $n{=}20$), $\lambda$ on a $19$-point grid over
$10^{-3}$--$10^{3}$, held-out RSS criterion, fixed fold shuffle.  EB-pc needs no tuning, but its
coefficient intervals are the least calibrated of the classical priors (Table~\ref{tab:t1coefcov}), so
we report both EB-iso and EB-pc throughout.

\paragraph{Elicitor provenance.}  Table~\ref{tab:provenance} lists
every model string and its main uses.

\begin{table}[h]\centering\small
\caption{Which model was used where.}\label{tab:provenance}
\resizebox{\columnwidth}{!}{%
\begin{tabular}{llll}
\toprule
Model string & Class & Role & Used in \\
\midrule
\texttt{claude-opus-4-7} & frontier (Opus) & primary, $K{=}3$ & E1--E16, R3, R17 \\
\texttt{gemini-3.1-flash-lite} & light flash & replication + & E16--E19, probe, \\
 & & all batteries & R22, $K{=}20$ \\
\texttt{gemini-3-flash-preview} & newer flash & post-hoc repl. & R17, R18, R26 \\
\bottomrule
\end{tabular}}
\end{table}

\paragraph{Training cutoffs.}  Vendor-documented cutoffs (model
documentation as read on 3 September 2026): \texttt{claude-opus-4-7},
reliable-knowledge and training-data cutoff January 2026;
\texttt{gemini-3.1-flash-lite} and \texttt{gemini-3-flash-preview},
knowledge cutoff January 2025 (stated for the Gemini~3 family).  All
NYC-2026 transactions (20 May--30 June 2026) and all recorded elicitation
timestamps (shipped with the transcripts) post-date these; the
Claude cutoff is the binding one, whereas the public benchmark
sources predate every cutoff by decades, so there the memorization
concern is addressed by anonymization and NYC-2026, not by dates.
Exposure does not imply recall either: models can fail to reproduce
their own fine-tuning data \citep{xue-etal-2026-supervised}.

\paragraph{Elicitation configuration (all batches).}  Anthropic
\texttt{claude-opus-4-7}: three recorded responses per prompt, stored
without call settings or timestamps.  Google
\texttt{gemini-3.1-flash-lite}: \texttt{generate\_content}
endpoint, temperature $0.7$ (E17 adds $0.2$, $1.0$).  Critical for the post-cutoff
logic, \emph{no tools, no search grounding and no retrieval
configuration of any kind were passed in any \gemini{} call}: the released
client code contains no grounding parameters.  Precisely:
no retrieval was \emph{requested} in any of these calls, and vendor
documentation states these endpoints do not retrieve by default;
client-side configuration cannot rule out undocumented server-side
behavior, so we state the guarantee at exactly this strength.
NYC-2026 data were
retrieved 2026-07-22 from NYC Open Data \citep{nycdof2026}, which states no restrictions
on use; all E18 elicitations
post-date that retrieval.  Every response ships verbatim, the \gemini{} responses with
timestamps, and the requests are regenerated by the released prompt
code; the probe's decision rule was
refined to a permutation-null test \emph{before} any probe
elicitation was made.

\begin{table}[htbp]\centering\scriptsize
\caption{Index of experiment codes: the question each code answers and where
it is reported (E $=$ original experiments; R $=$ first-revision analyses;
T1/T2 $=$ one-protocol re-evaluation and pre-specified batteries).}
\label{tab:index}
\setlength{\tabcolsep}{3pt}\renewcommand{\arraystretch}{0.75}
\begin{tabular}{@{}lp{5.2cm}l@{}}
\toprule
Code & Question & Where \\
\midrule
E1 & \raggedright Semi-synthetic California: gain, up to which $n$? & \S\ref{sec:E1}, Tab.~\ref{tab:rmse} \\
E1b & \raggedright Real California, same prior: loss-dependent ranking & \S\ref{sec:E1}, Tab.~\ref{tab:losses} \\
E2 & \raggedright Diabetes: best case; recall or knowledge? & \S\ref{sec:E1} \\
E4 & \raggedright Anonymization control (names removed) & \S\ref{sec:E4} \\
E5 & \raggedright Misaligned stress family: harm, and how reduced? & \S\ref{sec:E4} \\
E6, E12 & \raggedright Unknown $\sigma^2$; horseshoe & App.~\ref{app:addl} \\
E8 & \raggedright Log-pool vs.\ moment-match of the \claude{} draws & \S\ref{sec:E4} \\
E9 & \raggedright Logistic calibration, synthetic & App.~\ref{app:theoryexp} \\
E16, E17 & \raggedright Second vendor (Gemini); paraphrase, temperature & App.~\ref{app:crossmodel} \\
E13, E15 & \raggedright Breast Cancer; significance windows over $n$ & \S\ref{sec:E1} \\
E18 & \raggedright Post-cutoff NYC-2026; permuted-semantics probe & \S\ref{sec:E18} \\
E19 & \raggedright $K{=}20$ real draws; hierarchical pooling & \S\ref{sec:E4} \\
R1--R4 & \raggedright Center$\times$covariance factorial; gate; regret; coverage & \S\ref{sec:factorial} \\
T1 & \raggedright One protocol: four losses, oracle centers, $\tau_0^2$ & App.~\ref{app:t1} \\
T2 & \raggedright External test; permuted, paraphrase, no-default, derivable & App.~\ref{app:t2} \\
E11 & \raggedright Are elicitation \emph{errors} dependent? & \S\ref{sec:E4} \\
\bottomrule
\end{tabular}
\end{table}

\section{Reproducibility}\label{app:repro}
Code, recorded responses, data and results are at
\url{https://github.com/AprilZou852/llm-elicited-priors}.
All experiments run offline from the recorded elicitations on a single
CPU; four Tier-2 analyses need one external table that must be
downloaded.  \texttt{python code/run\_experiments.py} regenerates
E1--E10 and their figures, \texttt{code/run\_reviewer\_experiments.py}
the robustness studies, and the scripts listed in the repository's
README the rest.
Every number in Sections~\ref{sec:exp}--\ref{sec:disc} derives from
\texttt{T1\_*.csv} (Appendix~\ref{app:t1}), the processed R-tables
(Appendix~\ref{app:rtables}), the Tier-2 tables
(Appendix~\ref{app:t2}) or the E-series result files.  Prompts and
recorded responses are in \texttt{code/priors.py} and
\texttt{code/gap\_elicitations.json}; live re-elicitation
activates only when an API key is present.  The $51$ unit tests
(\texttt{pytest code/tests}) cover the conjugate updates, the
calibration fixed point and the empirical orderings reported above,
and \texttt{TABLE\_FIGURE\_MAP.csv} maps every printed table and
figure to its script and source file.

\section{Experiment Index}\label{app:index}
Table~\ref{tab:index} lists every experiment code used in the paper, the
question it answers and where its results are reported.

\end{document}